\documentclass[10pt,conference]{IEEEtran}
\IEEEoverridecommandlockouts
\usepackage{cite}
\usepackage{amsmath,amssymb,amsfonts}
\usepackage{graphicx}
\usepackage{textcomp}
\usepackage{xcolor}
\usepackage[numbers]{natbib}
\usepackage{textcomp}
\usepackage{xcolor}
\usepackage{tightenum}
\usepackage{placeins}
\usepackage{booktabs}
\usepackage{amssymb}
\usepackage{color}
\usepackage{xcolor}
\usepackage{subcaption}
\usepackage{graphics,graphicx,color}
\usepackage{multirow}
\usepackage{xspace}
\usepackage{algpseudocode}
\usepackage[ruled,vlined,linesnumbered]{algorithm2e}
\usepackage{enumitem}
\usepackage{ulem}
\usepackage{setspace}
\usepackage{url}

\newcommand{\xxx}{NSS\xspace}

\newcommand{\fdr}{fault detection rate\xspace}
\newcommand{\score}{TNSScore\xspace}

\usepackage[capitalize]{cleveref}
\crefname{section}{Sec.}{Secs.}
\Crefname{section}{Section}{Sections}
\Crefname{table}{Table}{Tables}
\crefname{table}{Tab.}{Tabs.}

\usepackage{tcolorbox}
\newcommand{\mybox}[1]{\begin{tcolorbox}[colback=yellow!10!white,colframe=red!75!black,lowerbox=invisible,savelowerto=\jobname_ex.tex]\emph{#1}\end{tcolorbox}}
\title{Neuron Sensitivity Guided Test Case Selection for Deep Learning Testing}

\begin{document}
\author{
	\IEEEauthorblockN{
		Dong Huang\textsuperscript{1}\textsuperscript{$\dagger$}\thanks{\textsuperscript{$\dagger$}These authors contributed equally to this work}, 
		Qingwen Bu\textsuperscript{2}\textsuperscript{3}\textsuperscript{$\dagger$}, 
        Yichao Fu\textsuperscript{1}, 
		Yuhao Qing\textsuperscript{1}, 
		Bocheng Xiao\textsuperscript{1} 
		Heming Cui\textsuperscript{1}\textsuperscript{3}
	\IEEEauthorblockA{\textsuperscript{1}The University of Hong Kong}
	\IEEEauthorblockA{\textsuperscript{2}Shanghai Jiao Tong University}
   \IEEEauthorblockA{\textsuperscript{3}Shanghai Artificial Intelligence Laboratory}
    \IEEEauthorblockA{\{dhuang, yhqing, heming\}@cs.hku.hk, qwbu01@sjtu.edu.cn, yichao@connect.hku.hk}
 }
} 
\maketitle

\begin{abstract}
Deep Neural Networks~(DNNs) have been widely deployed in software to address various tasks~(e.g., autonomous driving, medical diagnosis). However, they could also produce incorrect behaviors that result in financial losses and even threaten human safety. To reveal the incorrect behaviors in DNN and repair them, DNN developers often collect rich unlabeled datasets from the natural world and label them to test the DNN models. However, properly labeling a large number of unlabeled datasets is a highly expensive and time-consuming task. 

To address the above-mentioned problem, we propose \xxx, \textbf{N}euron \textbf{S}ensitivity guided test case \textbf{S}election, which can reduce the labeling time by selecting valuable test cases from unlabeled datasets. \xxx leverages the internal neuron's information induced by test cases to select valuable test cases, which have high confidence in causing the model to behave incorrectly. We evaluate \xxx with four widely used datasets and four well-designed DNN models compared to SOTA baseline methods. The results show that \xxx performs well in assessing the test cases' probability of fault triggering and model improvement capabilities. Specifically, compared with baseline approaches, \xxx obtains a higher fault detection rate~(e.g., when selecting 5\% test case from the unlabeled dataset in MNIST \& LeNe1 experiment, \xxx can obtain 81.8\% fault detection rate, 20\% higher than baselines).

\end{abstract}

\section{Introduction}\label{sec:intro}

Deep Neural Networks (DNNs) have become an increasingly important part of various tasks and are widely used to address a range of tasks, including autonomous driving~\cite{Bojarski2016EndTE}, medical diagnosis~\cite{Bakator2018DeepLA}, and machine translation~\cite{stahlberg2020neural}. The outstanding performance of DNN-driven software has revolutionized our daily lives. However, it has also brought attention to the quality and reliability of such DNN-driven software. Like conventional software, DNN-driven software is vulnerable to defects that can result in financial losses and threaten human safety~\cite{WinNT}. So, there is an urgent need for effective quality assurance techniques to be developed and applied to DNN-driven software to ensure their reliability and safety.

However, ensuring the quality of DNN-driven software is a complex issue, due to the differences between DNN models and conventional software systems~\cite{pei2017deepxplore, deepgini, gao2022adaptive}. Unlike traditional software systems that rely on developers' manual construction of business logic, DNNs are constructed based on a data-driven programming paradigm~\cite{deepgini,gao2022adaptive}. Thus, sufficient test data is critical for detecting and repairing incorrect behaviors of DNN-driven software~\cite{larochelle09a, CES, gao2022adaptive}, which requires DNN developers to collect a significant amount of data from various scenarios and hiring a large workforce to label it, which is a highly expensive and time-consuming task.

Under this situation, identifying and selecting the most valuable and representative data, which will be misclassified by DNN-driven software, becomes critical for improving the effectiveness and efficiency of quality assurance tasks for DNN-driven software~\cite{CES,gao2022adaptive, deepgini, kim2019guiding}. Inspired by the success of code coverage criteria in conventional software programs, prior researchers believe that test cases with higher neuron coverage results in higher adequacy and better quality of DNN testing. Then they proposed using neuron coverage to measure the adequacy of DNN testing~\cite{pei2017deepxplore,ma2018deepgauge, Gerasimou2020ImportanceDrivenDL}. For example, DeepXplore proposed NAC (Neuron Activation Coverage), which partitions neuron activation values into two ranges, each representing a state in the DNN. NAC then selects test cases that explore the most conditions in the DNN. These studies have demonstrated the effectiveness of neuron coverage in distinguishing general mutation tests. However, several studies~\cite{li2019structural, ismeanful, Yan2020CorrelationsBD} have revealed that higher neuron coverage does not correlate with a higher Fault Detection Rate (FDR, defined as the ratio of misclassified cases in selected cases), which limits their effectiveness as a guidance criterion for test case selection. Additionally, similar to conventional code coverage, neuron coverage also requires a high overhead in the collection process, making it difficult to apply to large-scale models~\cite{simonyan2014very} and datasets~\cite{deng2009imagenet}. 

To address these problems, recent researchers have proposed prioritization techniques that prioritize test cases based on some rules applied to the final layer outputs of DNNs~\cite{deepgini, CES, gao2022adaptive}. These techniques prioritize test cases with a high probability of detecting incorrect behaviors in DNNs, effectively collecting valuable test cases from a large unlabeled dataset. However, these prioritization techniques typically rely on the final layer outputs, which may not accurately reflect the internal neurons behaviors of the DNN and limit the visibility of the DNN developer~\cite{Ba2013DoDN, Papernot2015TheLO, Nguyen2019UnderstandingNN} since focusing on the final layer can not percept the global information of DNN. 
In many cases, DNN developers need to understand the internal logic of the DNN model and identify the root cause of incorrect behavior~(e.g., backdoor neurons~\cite{Liu2018TrojaningAO, Wang2019NeuralCI}) to effectively optimize and debug DNN-driven software~\cite{Xie2022NPCNP, Sun2018ConcolicTF}. Therefore, we believe that there is a need for a prioritization technique that utilizes DNN internal neuron information to help DNN developers efficiently debug incorrect behaviors.

Intuitively, the simplest way to exploit the internal neuron information is to extend existing prioritization techniques~(e.g., DeepGini~\cite{deepgini}, CES~\cite{CES}) to use the internal neuron output for prioritization directly. However, this was found to be impractical. 
Existing prioritization techniques leverage the confidence output with respect to each label in the final layer of the models. Therefore, internal neurons cannot be utilized directly for prioritization since they do not carry this semantic information.

\begin{figure}[t]
    \centering
    \includegraphics[width=\linewidth]{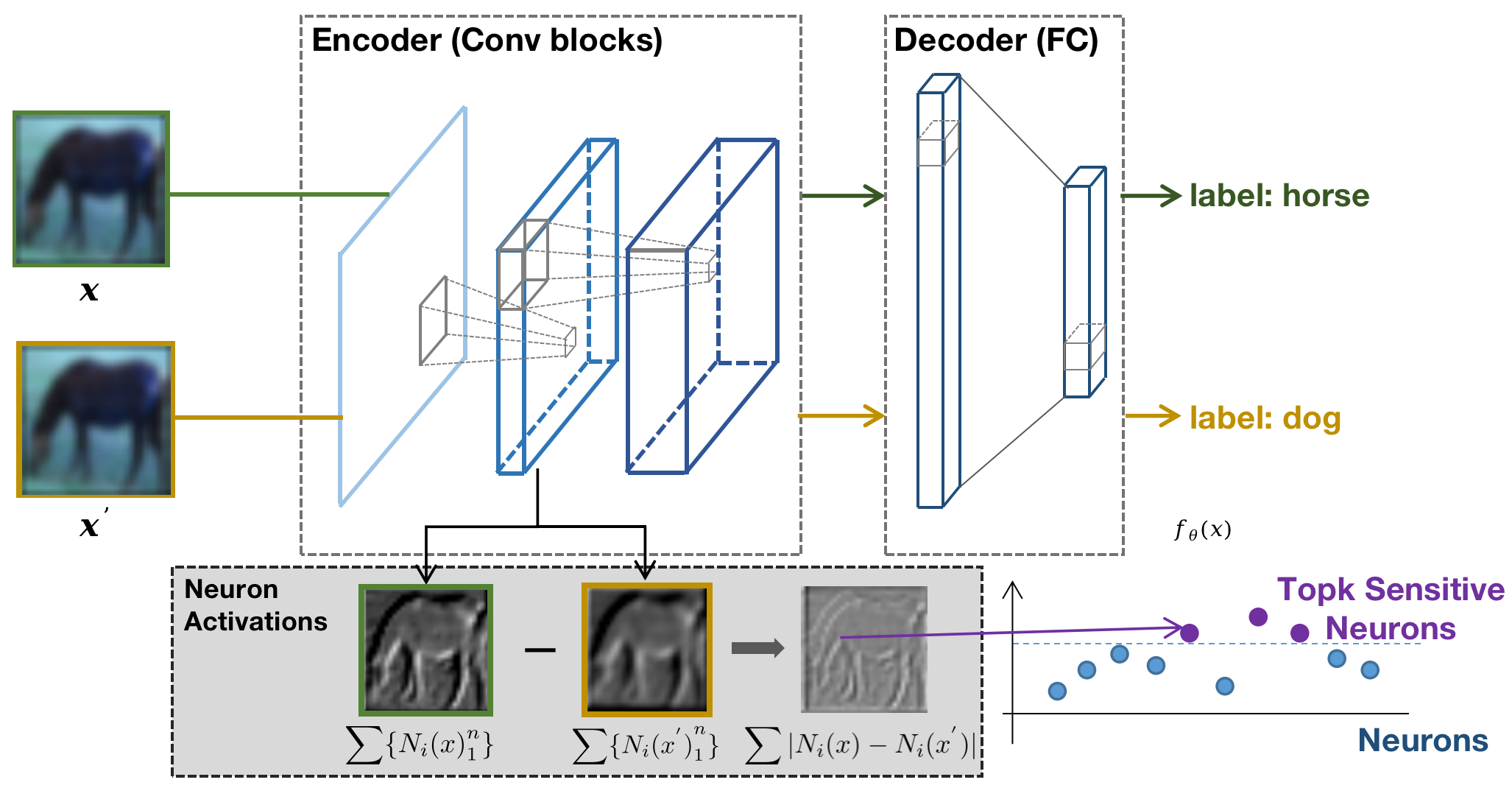}
    \caption{\textit{Neurons Sensitivity} is measured by the difference between the neuron activation values within the network, when the inputs are clean and mutated samples, respectively.} 
    \label{fig:intro}
    \vspace{-0.7cm}
\end{figure}

To address this challenge, we propose \xxx~(i.e., \textbf{N}euron \textbf{S}ensitivity guided test case \textbf{S}election), which is inspired by the concept of "neuron sensitivity" in the deep neural network~(see~\cref{fig:intro}). 
Neuron sensitivity, which measures the impact of input perturbations on the model's output, is correlated with DNN performance \cite{Zhang2019InterpretingAI, Pizarroso2020NeuralSensSA}. Specifically, we observe that a test case with high sensitivity values for a neuron is more likely to cause the model to produce incorrect behavior. To prioritize test cases with neuron sensitivity, we propose a novel Test case's Neuron Sensitivity Score (TNSScore), which represents the sum of a test case's neuron sensitivity values across all neurons in the DNN. A test case with a higher TNSScore means it has higher confidence to detect incorrect behavior in DNN. However, since DNNs can have millions of neurons, calculating the TNSScore for all neurons can be computationally expensive. To address this issue and improve efficiency,  we propose a Sensitive Neuron Identifier that detects sensitive neurons in the DNN using a subset of the unlabeled dataset. By identifying the sensitive neurons, we can reduce the computation time required to calculate the TNSScore, as we only need to calculate the TNSScore for the sensitive neurons. Thus our approach enables the effective and efficient detection of valuable test cases with high confidence in detecting incorrect behavior in DNNs.

To validate the effectiveness of \xxx, we conduct experiments with four well-designed DNN models across four widely-used datasets. We also use seven widely-used data mutation strategies to generate unlabeled candidate datasets in our experiments. The experiment results demonstrate that \xxx performs well in test case selection tasks. Specifically, compared with baseline approaches, \xxx obtains higher fault detection rate~(FDR)~(e.g., when selecting 5\% test case from the unlabeled dataset in MNIST \& LeNet1 combination, \xxx can obtain 81.8\% FDR, which increases 20\% FDR compared with baseline approaches). Employing the selected test cases to retrain models can increase accuracy more than baselines.

In a nutshell, we make the following contributions.
\begin{itemize}

    \item We define the Test case Neuron Sensitivity Score~(TNSScore), which could measure the confidence of a test case misclassified by the DNN model.
    Then, we propose neuron sensitivity guided test case selection~(\xxx), which could select valuable test cases from unlabeled datasets.
    
    \item We conduct extensive experiments to investigate the performance of \xxx. The results show that \xxx can significantly outperform other test selection methods and efficiently enhances the DNN model.
    
    \item We implement \xxx into a tool that could help developers to select test cases from massive unlabelled datasets, which is available in our GitHub Page~\cite{sourcecode}. 
    
\end{itemize}
\section{Background}\label{sec:back}

\subsection{Neural Network}
In our work, we focus on Deep Neural Networks~(DNNs) for classification, which can be presented as a complicated function $f: \mathcal{X} \rightarrow \mathcal{Y}$ mapping an input $x \in \mathcal{X}$ into a label $y\in\mathcal{Y}$. Unlike traditional software, programmed with deterministic algorithms by developers, DNNs are defined by the training data, along with the network structures. Generally speaking, a DNN model consists of an input layer, an output layer, and at least one hidden layer. Each neuron in each layer is intertwined with neurons in other layers, and the output of each neuron is the weighted sum of the outputs of all neurons in the previous layer. Then a nonlinear activation function (e.g.,tanh, sigmoid, and ReLU) is applied.



\subsection{Neuron Sensitivity}
The neuron sensitivity is widely used in the AI community to represent the specific neuron's  behaviors~\cite{Zhang2019InterpretingAI}. 
Formally, the neuron sensitivity $S(N_i, (x,x'))$ of a neuron $N_i$ in a DNN model over a dual pair set $(x,x')$ is defined as the average $L_1$ norm of the differences between the neuron outputs on the pairs of inputs, normalized by the dimension of the neuron output vector, i.e., 
$$S(N_i, (x,x'))= \left|N_i\left(x\right)-N_i\left(x^{\prime}\right)\right|$$
where $N_{i}(x)$ is the output of neuron $N_{i}$ on input $x$, $\left| \cdot \right|$ denotes the $L_1$ norm, and the $x'$ is mutated by $x$ with DNN developers specified mutation strategies~(e.g., rotation, shear, blur).


\subsection{Neuron Coverage Metrics}

\paragraph{\textbf{Neuron Activation Coverage~(NAC($k$))}}
NAC($k$) was proposed by DeepXplore~\cite{pei2017deepxplore}, NAC($k$) assumes that the more neurons are activated, the more states of DNN are explored. The developer defines the parameter $k$ of this coverage criteria to specify how a neuron in a DNN can be counted as covered.

\paragraph{\textbf{K-multisection Neuron Coverage~(KMNC($k$))}}
Based on the NAC($k$) assumption about the DNN states, DeepGuage~\cite{ma2018deepgauge} further partitions the neuron's output into $k$ ranges, and each range represents one state in DNN. 





\subsection{Test Case Selection Methods}
In this section, we introduce several widely used test case selection methods, which are used to select valuable test cases from a massive number of datasets. Specifically, the goal of the test case selection method is to sampler a fixed size~($N$) subset $I_{N}$ from the total test set $I_{T}$. We divide test case selection methods into two types: coverage-guided test case selection and prioritization test case selection.
\subsubsection{Coverage-guided test case selection}
Coverage-guided selection methods~(e.g., NAC, KMNC, and NPC) try to select test cases that can reach maximum coverage metrics and lead to a higher fault detection rate~\cite{pei2017deepxplore,kim2019guiding,ma2018deepgauge, Xie2022NPCNP}.

\subsubsection{Prioritization test case selection}
Generally, for a given total test set $I_{T}$, prioritization test selection methods~(e.g., DeepGini~\cite{deepgini}, SA~\cite{kim2019guiding}, ATS~\cite{gao2022adaptive}) compute a probability $p_{i}$ for each test case $i$ in the total test set. The value of $p_{i}$ represents the probability of test case $i$ sampled by selection methods. 

\begin{table}[]
    \small
    \setlength{\tabcolsep}{2.5pt}
    \centering
    \begin{tabular}{ccccc}
    \toprule
    Dataset&DNN Model &Neurons & Layers&Ori Acc (\%)\\
    \midrule
    \multirow{2}*{MNIST~\cite{deng2012mnist}} & LeNet-1~(L-1)~\cite{lecun1998gradient}&3,350&5&89.50\\
    ~&LeNet-5~(L-5)~\cite{lecun1998gradient}&44,426&7&91.79\\
    \midrule
    \multirow{2}*{CIFAR-10~\cite{Krizhevsky09learningmultiple}}& ResNet-20~(R-20)~\cite{he2016deep}&543,754&20&86.07\\
    ~&VGG-16~(V-16)~\cite{simonyan2014very}&35,749,834&21&82.52\\
    \midrule
    \multirow{2}*{Fashion~\cite{xiao2017/online}}& LeNet-1~(L-1)~\cite{lecun1998gradient}&3,350&5&78.99\\
    ~&ResNet-20~(R-20)~\cite{he2016deep}&543,754&20&86.12\\
    \midrule
    \multirow{2}*{SVHN~\cite{Netzer2011}} & LeNet-5~(L-5)~\cite{lecun1998gradient}&44,426&7&84.17\\
    ~&VGG-16~(V-16)~\cite{simonyan2014very}&35,749,834&21&92.02\\
    \bottomrule
    \end{tabular}
    \caption{Datasets and DNNs for evaluating \xxx, which covers the complete set of datasets evaluated by baselines.}
    \vspace{-0.7cm}
    \label{tab:dataset}
\end{table}
\section{Methodology}\label{sec:overview}

\begin{figure}
    \centering
    \includegraphics[width=0.8\linewidth]{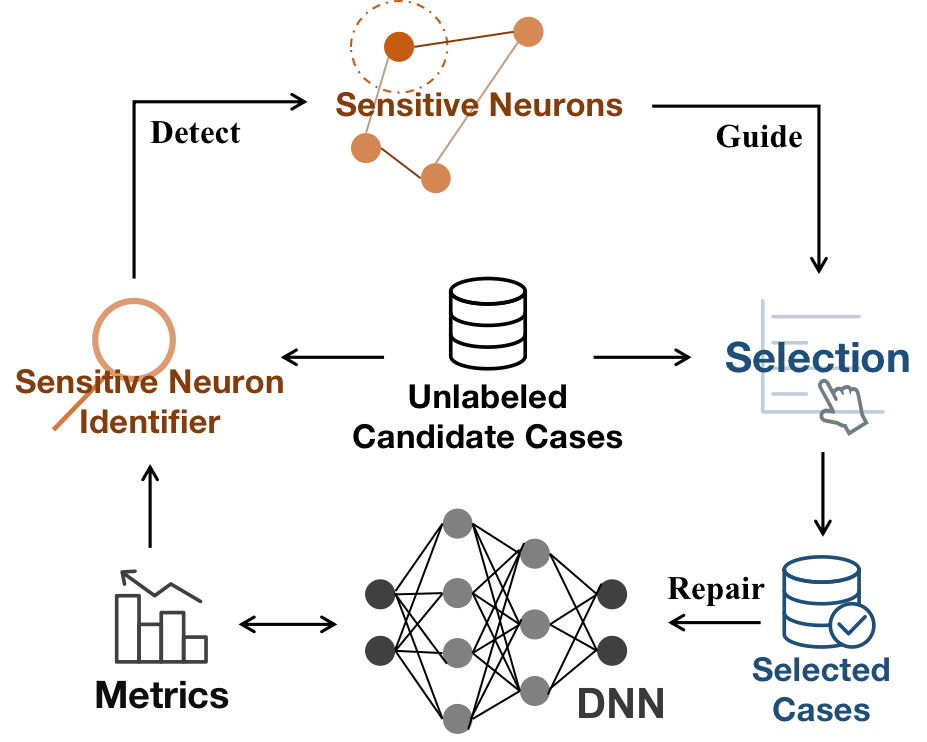}
    \caption{Overview of \xxx's workflow.}
    \label{fig:pipeline}
    \vspace{-0.7cm}
\end{figure}

The overview of \xxx's workflow is shown in~\cref{fig:pipeline}, which mainly includes two stages: 1. detect sensitive neurons based on neuron sensitivity and 2. test case selection from unlabeled dataset~(i.e., prioritize test case by TNSScore, which is defined in~\cref{sec:method:metric}). Specifically, during the testing process~(\cref{sec:method:identifier}), \xxx first feeds unlabeled candidate dataset into \textit{sensitive neuron identifier} to detect sensitive neurons in the DNN model, which can reduce the computational expense, and then it reports sensitive neurons to DNN developers.
Then unlabeled candidate dataset will be fed into the model to calculate the Metric Score on the sensitive neurons, i.e., the Test case's Neuron Sensitivity Score~(\textbf{TNSScore}). 
Next, \xxx prioritizes test cases based on their \score~(\cref{sec:selection}).
Finally, the cases with high Scores will be selected by \xxx.

For ease of discussion, this section defines the following notations for DNNs and neurons: $f_\theta$ is a DNN parameterized by $\theta$, and there are $n$ neurons~(i.e., $\{N_{i}\}_{1}^{n}$) in the model. The $i$-th neuron in the DNN is denoted as $N_{i}$. $N_{i}(x)$ denotes the output of the corresponding neuron when the network input is $x$. The mutation case x' is generated by benign mutation strategies~(e.g., rotation, blur, scale).

\subsection{Test case's Neuron Sensitivity Score}\label{sec:method:metric}

Neuron sensitivity provides valuable insights into the behavior of neurons within a DNN model and helps assess the reliability of a neuron~\cite{Zhang2019InterpretingAI}. To better prioritize test cases based on their impact on the DNN, we propose the Test case's Neuron Sensitivity Score (TNSScore). The TNSScore measures the model's sensitivity to a given test case and helps evaluate the likelihood of the test case being misclassified.

Formally, given a test case $x$ and a set of internal neurons ${N_{i}}_{1}^{n}$ in a DNN, the TNSScore of test case $x$ is defined as:

$$TNSScore_{x} = \sum_{i=1}^{n}|N_{i}(x) - N_{i}(x')| = \sum_{i=1}^{n}S(N_{i},(x,x'))$$

To calculate the TNSScore, \xxx first feeds the candidate test cases into the DNN to obtain the activation values for each neuron. Next, a benign mutation strategy (e.g., rotation, blur, scale) is applied to each test case to generate a mutated version $x'$. Finally, \xxx calculates the difference between the activation values of each neuron for the original and mutated test case and uses this information to obtain the TNSScore for each test case. This metric allows developers to prioritize test cases based on their TNSScore values. Our experiments have shown that test cases with higher TNSScores are more likely to detect incorrect behavior in the DNN model.

\subsection{Sensitive Neuron Identifier}\label{sec:method:identifier}
Intuitively, prioritizing test cases based on the TNSScore value in all neurons is convincing. However, we notice that in the DNN model, the number of internal neurons is enormous. For example, the VGG16 model has 35,749,834 neurons, and the ResNet20 model, widely used in deep learning tasks, has 543,754 neurons, so calculating the TNSScore value in all neurons is unrealistic because it is a highly expensive and time-consuming task.  

To address this problem, we propose \textit{Sensitive Neuron Identifier}~(Identifier), which is designed to identify a subset of the most sensitive neurons in a given deep neural network. Using these neurons to calculate the \score can reduce the calculating time in the process.
The detailed implementation of the \textit{Sensitive Neuron Identifier} is provided in Alg~\ref{algo:identifier}. Specifically, given a dataset of input samples, we randomly produce a set of samples from the dataset~(line 5). Next, we compute the sensitivity of neurons with respect to the selected sample pairs and record the most sensitive neurons to each pair~(lines 6-9). Finally, we select the neurons found to be sensitive in most sample pairs as the final result~(line 10). 

\begin{algorithm}
    \caption{Sensitive Neuron Identifier}\label{algo:identifier}
    \SetKwInput{Vars}{Variables}
    \SetKwProg{Fn}{Function}{:}{}
    \DontPrintSemicolon
    \SetNoFillComment
    \SetKwFunction{Fun}{Fun}
    \SetKwInOut{KOutput}{output}
    \KwIn{
     $x\in\mathcal{X}$: original test cases;
     $f_{\theta}$: DNN to be tested;
     $k$: Percentage of sensitive neuron.
     }
     \KOutput{$\mathcal{X_{s}}$: selected test cases}

    \Fn{Identifier($f_{\theta},\mathcal{X}, k$)}
    {Initilize NSList = [0]*n: Initialize Neuron Sensitivity List;\\
    Initialize SNIdx: initialize sensitive neuron index list;\\
    \For{$x \in \mathcal{X}$}{
    $x' = BenignMutation(x)$;\
    $\{{N_{i}(x)}\}_{i=1}^{n} \leftarrow f_{\theta}(x)$'s activtion values;\
    $\{{N_{i}(x')}\}_{i=1}^{n} \leftarrow f_{\theta}(x')$'s activtion values;\
    \For {$i \in $ [1,...,n] } {
    $NSList$[i]+=($|N_{i}(x') - N_{i}(x)|$)\;
    }
    }
    SNIdx = $ArgSort(NSList)[-k*len(NSList):]$ \
    }
    \Return SNIdx
\end{algorithm}

\begin{algorithm}
    \caption{\xxx Test Case Selection}\label{algo:selection}
    \SetKwInput{Vars}{Variables}
    \SetKwProg{Fn}{Function}{:}{}
    \DontPrintSemicolon
    \SetNoFillComment
    \SetKwFunction{Fun}{Fun}
    \SetKwInOut{KOutput}{output}
    \KwIn{
     $x\in\mathcal{X}$: original test cases;
     $f_{\theta}$: DNN to be tested;
     $k$: Percentage of sensitve neuron detected by Sensitive Neuron Identifier;
     $N$: The number of test cases to be selected by \xxx.\
     }
     \KOutput{$\mathcal{X_{s}}$: selected test cases}

     \Fn{\xxx($f_{\theta}, \mathcal{X}, k, N$)}
     {
    All\_TNSScore = []\
     \tcc{Detect sensitive neuron in DNN}\
    SNIdx = Identifier($f_{\theta}, \mathcal{X}, k$)\
    \tcc{Prioritize Test Case based on TNSScore}\
    
    \For{$x \in \mathcal{X}$}{
    TNSScore = 0\
    $x' = BenignMutation(x)$\
    $\{{N_{i}(x)}\}_{i=1}^{n} \leftarrow f_{\theta}(x)$'s activtion values \; 
    $\{{N_{i}(x')}\}_{i=1}^{n} \leftarrow f_{\theta}(x')$'s activtion values \; 
    \For {$i \in $ SNIdx } {
    TNSScore+= $|N_{i}(x') - N_{i}(x)|$\;
    }
    All\_TNSScore.append(TNSScore)
    }
    $\mathcal{X_{s}} = \mathcal{X}[argsort($All\_TNSScore$)[-N:]]$\\
    \Return $\mathcal{X_{s}}$ \
    }
\end{algorithm}

\subsection{Prioritizing Test Case by Neuron Sensitivity}\label{sec:selection}
Both testing and repairing the DNN-driven system rely on manually labeled data. While collecting a massive amount of unlabeled data is usually easy to achieve, manual labeling costs much greater. For data with strong expertise knowledge~(e.g., medical data), it is unrealistic to label all collected data blindly. Therefore, DNN test case selection is crucial to select valuable data, reducing the labeling cost.
To select test cases with a high fault detection rate, we propose a neuron sensitivity-guided test case selection. Alg~\ref{algo:selection} shows the workflow of our selection. It first utilizes \textit{ Sensitive Neuron Identifier} to detect sensitive neurons in DNN~(line 3). Then it computes the TNSScore in the sensitive neuron between each original test case $x$ and its corresponding mutation case $x'$~(lines 4-11). Then the test case will be prioritized by their score, i.e., the test cases with a higher score will be selected by \xxx~(lines 12-13).

\paragraph{Example} Here, we use a simplified example to illustrate how \xxx selects test cases from unlabeled datasets. Specifically, assume that we have four different test case $x_{1}$, $x_{2}$, $x_{3}$, and $x_{4}$, and the DNN have two sensitive neuron $N_{1}$, and $N_{2}$. We first use benign mutation in the test cases to generate its corresponding mutation cases $x_{1}'$, $x_{2}'$, $x_{3}'$, and $x_{4}'$. Then we feed these cases into the model to obtain the neuron's activation output and neuron sensitivity. According to the value of $S$ in ~\cref{tab:example_prioritize_tests}, we can prioritize the tests as $x_{3}$, $x_{1}$, $x_{4}$, and $x_{2}$. Thus, DNN is the most sensitive for $x_{3}$'s change, which indicates that DNN lacks sufficient background knowledge about $x_{3}$ and needs $x_{3}$ related knowledge to enhance the DNN. 
While for $x_{2}$, DNN's neuron will not be affected by the $x_{2}$'s mutation change, which indicates that the DNN model has sufficient knowledge about $x_{2}$, so that $x_{2}$ is not a valuable case for DNN now~(i.e., using $x_{2}$ to repair DNN will not change DNN's parameter).

\begin{table}[]
\small
    \centering
    \begin{tabular}{c|c |c| c}
    \toprule
        Tests $x,x'$&$N_{1}(x),N_{1}(x'),S_{1}$&$N_{2}(x),N_{2}(x'),S_{2}$&$S$\\
        \midrule
         $x_{1}$,$x_{1}'$&0.4, 0.3, 0.1&0.5, 0.4, 0.1&0.2 \\
         $x_{2}$,$x_{2}'$&0.2, 0.2, 0&0.4, 0.4, 0&0 \\
         $x_{3}$,$x_{3}'$&0.4, 0.3, 0.1&0.5, 0.3, 0.2&0.3 \\
         $x_{4}$,$x_{4}'$& 0.8, 0.7, 0.1&0.5, 0.45, 0.05&0.15\\
         \bottomrule
    \end{tabular}
    \caption{An example to show how \xxx prioritize test cases.}
    \label{tab:example_prioritize_tests}
    \vspace{-0.3cm}
\end{table}

\section{Evaluation}

We evaluate \xxx and answer the following questions.

\begin{itemize}
    \item RQ1~(Sensitivity): What is the correlation between neuron sensitivity and model performance?
    \item RQ2~(Selection): How effective and efficient is \xxx?
    \item RQ3~(Sample Size): How does sensitive neuron sample size affect \xxx's effectiveness?
    \item RQ4~(Layer Selection): How does the selected layer affect \xxx's effectiveness?
\end{itemize}

\subsection{Experiment Setup}
Our evaluation was done on a GPU server with two twenty-core CPUs and four NVIDIA RTX 2080Ti graphic cards. 

\begin{table}[t]
\small
    \centering

    \begin{tabular}{c|c c}
    \toprule
         Transformations&Parameters& Parameter ranges  \\
    \midrule
         Shift&$(s_x, s_y)$ &[0.05, 0.15]\\
         Rotation&$q~(degree)$&[5, 25]\\
         Scale&$r~(ratio)$&[0.8,1.2]\\
         Shear&$s~(angle)$&[15, 30]\\
         contrast&$\alpha~(gain)$&[0.5,1.5]\\
         Brightness& $\beta~(bias)$&[0.5,1.5]\\
         Blur&$ks~(kernel size)$&\{2,3,5,7\}\\
    \bottomrule
    \end{tabular}
    \caption{Transformations and parameters used by \xxx for generating unlabeled test cases.}
    \label{tab:setup}
    \vspace{-0.7cm}
\end{table}

\begin{table}[]
\small
    \centering
    \begin{tabular}{c|c c}
    \toprule
         Criteria&Parameters &Parameter Config \\
         \midrule
         Random&- &-\\
         NAC&t~(threshold)&0.5\\
         KMNC&k~(k-bins)&1000\\
         NPC&$\alpha$&0.7\\
         DSA&$n$&1000\\
         Gini&None&None\\
         ATS&None&None\\
         \xxx&k~(percentage of sensitive neuron)&10(\%)\\
    \bottomrule
    \end{tabular}
    \caption{The parameter configuration of test case selection.}
    \label{tab:config}
    \vspace{-0.7cm}
\end{table}

\paragraph{\textbf{Datasets and Models.}}
We adopt four widely used image classification benchmark datasets for the evaluation (i.e., MNIST~\cite{deng2012mnist}, Fashion MNIST~\cite{xiao2017/online}, SVHN~\cite{Netzer2011ReadingDI}, and CIFAR10~\cite{Krizhevsky09learningmultiple}), which are most commonly used datasets in deep learning testing~\cite{pei2017deepxplore,DeepMutation++,deepgini,ma2018deepgauge,Xie2018DeepHunterHD,Tian2018DeepTestAT,kim2019guiding,Gerasimou2020ImportanceDrivenDL,ismeanful,ADAPT,gao2022adaptive,Wang2021RobOTRT}. Table \ref{tab:dataset} presents the detail of the datasets and models. The MNIST~\cite{deng2012mnist} dataset is a large collection of handwritten digits. It contains a training set of 60,000 examples and a test set of 10,000 examples. The CIFAR-10~\cite{Krizhevsky09learningmultiple} dataset consists of 60,000 32x32 color images in 10 classes, with 6,000 images per class. Fashion~\cite{xiao2017/online} is a dataset of Zalando's article images—consisting of a training set of 60,000 examples and a test set of 10,000 examples. SVHN~\cite{Netzer2011} is a real-world image dataset that can be seen as similar in flavor to MNIST (e.g., the images are of small cropped digits). The models we evaluated include LeNet~\cite{lecun1998gradient}, VGG~\cite{simonyan2014very}, and ResNet~\cite{he2016deep}, which are also commonly used in deep learning testing tasks~~\cite{pei2017deepxplore,DeepMutation++,deepgini,ma2018deepgauge,Xie2018DeepHunterHD,Tian2018DeepTestAT,kim2019guiding,Gerasimou2020ImportanceDrivenDL,ismeanful,ADAPT,gao2022adaptive,Wang2021RobOTRT}. 

\paragraph{\textbf{Test Case Generation.}}
We follow the prior benign data simulation~\cite{Tian2018DeepTestAT,Ma2018DeepMutationMT, gao2022adaptive} strategies to generate realistic unlabeled datasets. Specifically, we use seven widely-used benign mutations~(i.e., shift, rotation, scale, shear, contrast, brightness, and blur) to generate the test case with its original label. The parameters of the mutation are shown in~\cref{tab:setup}. We do not choose adversarial attack~(e.g., FGSM, PGD, and BIM) to generate test cases because these data can not represent data collected from the real-world scenario and may lead to unreliable conclusions~\cite{li2019structural}.
During the test case generation, for each test data in the dataset, we randomly select one benign data augmentation from our seven augmentations to mutate a test case with its original label. 
When the original test size is 10,000, we will generate the test cases with the same size.

\paragraph{\textbf{Test Case Selection.}}
Multiple test case selections have been proposed by recent works~(e.g., NC-guided~\cite{pei2017deepxplore,Tian2018DeepTestAT,ma2018deepgauge,Gerasimou2020ImportanceDrivenDL,Xie2022NPCNP}, Priority~\cite{deepgini,Wang2021PrioritizingTI}, Active learning guided~\cite{gao2022adaptive,Ren2020ASO}, robust-guided~\cite{Wang2021RobOTRT,madry2017towards, Chen2020PracticalAE}, SA-guided~\cite{kim2019guiding}).
However, robust-guided selection~\cite{Wang2021RobOTRT,madry2017towards,Chen2020PracticalAE} focus on model robustness, as mentioned by~\citet{Zhang2019TheoreticallyPT}, there is a trade-off between the accuracy and robustness, which means using these~(e.g., RobOT~\cite{Wang2021RobOTRT}, and PACE~\cite{Chen2020PracticalAE}) strategies to increase model robustness, the accuracy will decrease. So we will not choose these strategies as our baselines. 

To show the effectiveness of neuron sensitivity-guided test case selection, we use the most famous metric~(i.e., NAC), K-multisection neuron coverage~(i.e., KMNC) proposed by DeepGauge~\cite{ma2018deepgauge} and Neuron Path Coverage~(NPC), which is used to evaluate the covered path in DNN decision flow. Compared with other works~(e.g., IDC, MC, and DC) that select test cases to induce one neuron behavior, DeepGauge can select test cases to induce multiple neuron behaviors. Since \xxx is a prioritization technique~(i.e., priority the test cases with certain rules and return cases with larger priority), we take DeepGini, the SOTA open-sourced prioritization technique as our baseline. To compare \xxx with SA metrics, we use DSA, which is proposed by~\citet{kim2019guiding}, to evaluate \xxx's effectiveness. Then to evaluate \xxx's effectivenss with the current SOTA active learning guided selection strategies, we use ATS~\cite{gao2022adaptive} as our baseline. Finally, we also use Random Selection~(RS) as a natural baseline, which can help us evaluate whether a selection method is effective. 
The parameters of the selection strategies are shown in~\cref{tab:config}.


\begin{figure*}[]
	\centering
	\begin{subfigure}{0.22\linewidth}
		\centering
		\includegraphics[width=1\linewidth]{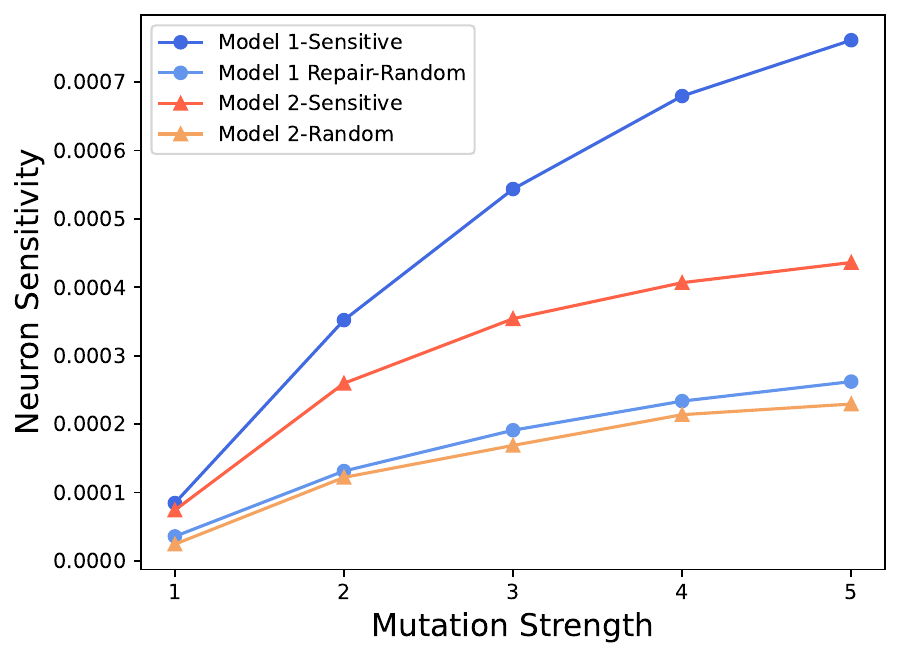}
		\caption{Fashion \& ResNet20}
		\label{chutian3}
	\end{subfigure}
		\begin{subfigure}{0.24\linewidth}
		\centering
		\includegraphics[width=1\linewidth]{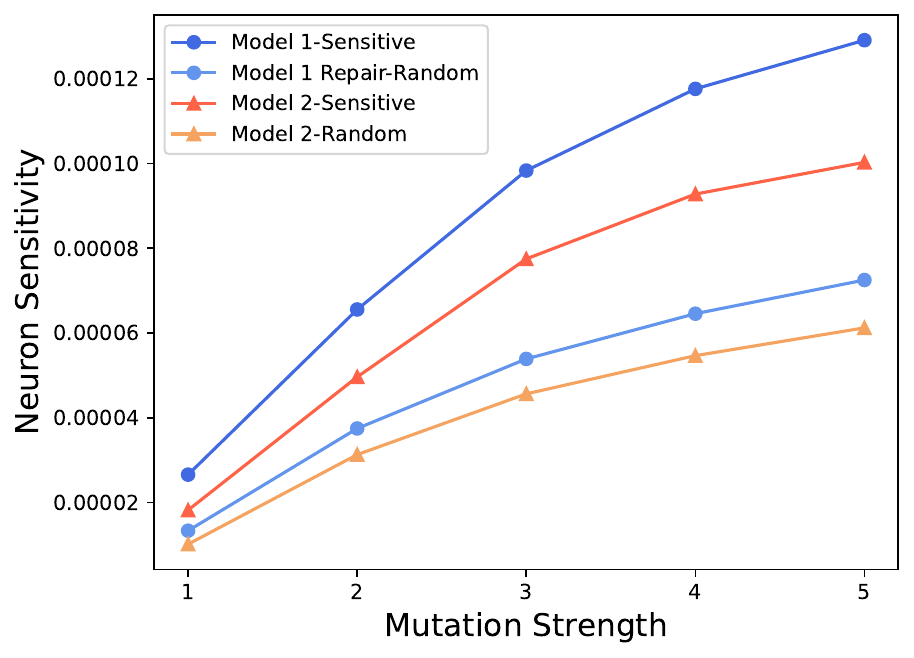}
		\caption{CIFAR10 \& ResNet20}
		\label{chutian3}
	\end{subfigure}
	\centering
	\begin{subfigure}{0.24\linewidth}
		\centering
		\includegraphics[width=1\linewidth]{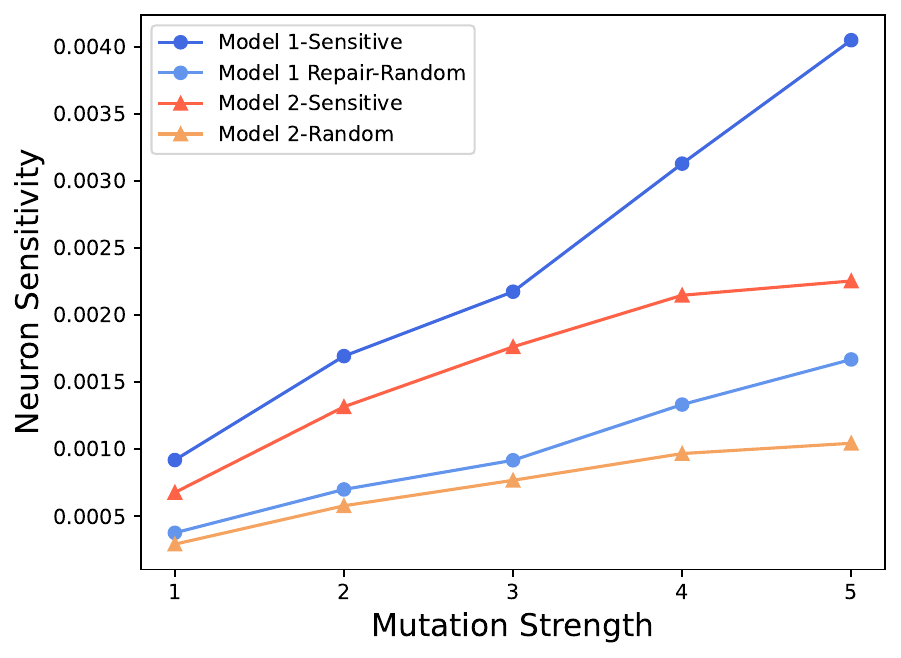}
		\caption{CIFAR10 \& VGG16}
		\label{chutian3}
	\end{subfigure}
	\centering
	\begin{subfigure}{0.24\linewidth}
		\centering
		\includegraphics[width=1\linewidth]{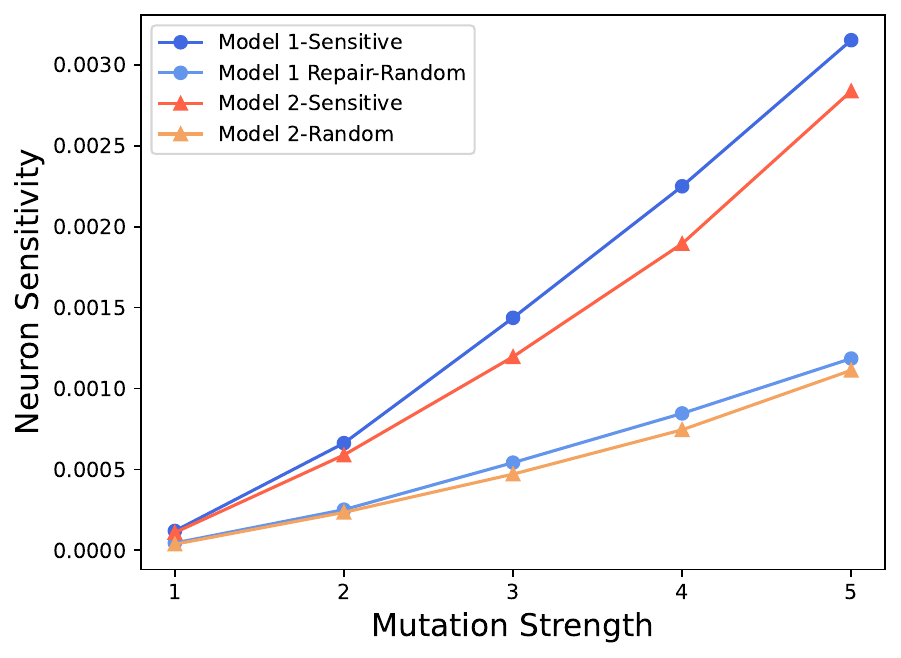}
		\caption{SVHN \& VGG16}
		\label{chutian3}
	\end{subfigure}
 \vspace{-0.2cm}
	\caption{The neuron sensitivity of the Top 10\% sensitive and randomly selected neurons under different mutation strength and model performance, where \textit{model 1} denotes the model with relatively lower accuracy. We only report four combinations in our paper due to page limitation, and other combinations exhibit similar experiment results.}
	\label{fig:sensitivity}
 \vspace{-0.3cm}
\end{figure*}

\subsection{What is the correlation between neuron sensitivity and model performance?}\label{sec:eval:sensitivity}
To validate the effectiveness and correctness of our metric, we first evaluate the correlation between neuron sensitivity and model error rate on eight different models \& dataset combinations. For each model, we train two different accuracy stages and we analyze the neurons' sensitivity in different accuracy stages of the model. Results are shown in~\cref{fig:sensitivity}, where \textit{model 1} denotes the model with relatively lower accuracy. 
We can observe that for each different model in our experiments, the model with higher error rates on the given dataset exhibit higher neuron sensitivity across all dataset \& model combinations. Additionally, with the increase of mutation strength, sensitive neurons' average sensitivity rises in a manner consistent with that of all other neurons at the same time, while exhibiting a greater range of variation. Differences in absolute values and magnitude of change, between sensitive neurons and randomly selected neurons, indicate that there are indeed significantly more sensitive neurons in the neural network. 
\mybox{Answer to RQ1: Neuron sensitivity is strongly correlated with model performance. Models with a higher error rate will have higher neuron sensitivity.}

\begin{table*}[t]
    \centering
    \setlength{\tabcolsep}{1.8pt}
        \begin{tabular}{l|c|ccccccc|c|ccccccc}
        \toprule
        \multirow{2}{*}{Dataset(DNN)}&\multicolumn{8}{c|}{Select 5\% Test Cases}&\multicolumn{8}{c}{Select 10\% Test Cases}\\
        &\xxx&NAC&KMNC&NPC&DSA&Gini&ATS&RS&\xxx&NAC&KMNC&NPC&DSA&Gini&ATS&RS\\
        \midrule
        MNIST~(L-1)&{\color{green}81.8}&22.6 &42.2 &20.2 &22.2 &61.8&58.7&21.3&      {\color{green}70.6}&20.8 &38.7&20.2 &22.5&55.7&54.3&21.3\\
        MNIST~(L-5)&{\color{green}81.8}& 18.4& 25.8& 21.6& 22.6&58.2&60.2&18.8&      {\color{green}67.7}& 18.1& 20.8& 20.0& 21.3&50.5&52.3&18.8\\
        Fashion~(L-1)&{\color{green}70.2}& 32.7& 35.9& 31.0& 31.1&57.8&53.3&31.2&    {\color{green}65.0}& 30.9& 35.9& 31.0& 31.0&48.2&48.7&31.0\\
        Fashion~(R-20)&{\color{green}89.4}& 21.1& 23.7& 30.5& 24.1&55.0&40.3&26.3&    {\color{green}80.7}& 20.5& 22.0& 28.4& 27.1&48.7&35.2&26.2\\
        SVHN~(L-5)&52.7& 29.1& 28.8& 31.0& 31.1&{\color{green}53.2}&47.8&28.8&       {\color{green}52.5} & 29.1 & 27.2 & 31.0 & 31.1&47.3&42.1&29.1\\
        SVHN~(V-16)&{\color{green}77.3}&21.5 &16.2 &18.9 &24.5 &53.0&55.3&16.0&        {\color{green}63.8}& 19.6&16.2 & 17.1&23.4&43.1 &48.7&16.0 \\
        CIFAR-10~(V-16)&{\color{green}68.6} & 28.4 &  16.4&30.6  &43.6 &60.2&62.1&30.9&{\color{green}64.8}& 29.4& 17.4&30.3 & 43.3&56.2&56.3&30.8\\
        CIFAR-10~(R-20)&{\color{green}71.4}& 28.2& 24.6& 26.0& 27.4&50.4&50.2&28.8& {\color{green}62.8}& 25.8& 21.8& 28.7& 27.4&45.0&47.1&28.9\\
        \midrule
        \multirow{2}{*}{Dataset(DNN)}&\multicolumn{8}{c|}{Select 15\% Test Cases}&\multicolumn{8}{c}{Select 20\% Test Cases}\\
        &\xxx&NAC&KMNC&NPC&DSA&Gini&ATS&RS&\xxx&NAC&KMNC&NPC&DSA&Gini&ATS&RS\\
        \midrule
        MNIST~(L-1)&{\color{green}58.3}& 21.6& 35.3& 20.6& 23.6&50.1&47.6&21.3& {\color{green}53.3} & 22.0 & 35.7 & 20.9 & 23.5&45.4&41.5&21.3\\
        MNIST~(L-5)&{\color{green}56.7}& 18.5& 19.3& 19.3& 21.0&46.2&40.2&18.7&{\color{green}50.8} & 18.4 & 19.4 & 19.1 & 20.8&42.3&37.1&18.7\\
        Fashion~(L-1)&{\color{green}59.6} & 30.5 & 34.2 & 31.0 & 31.0&44.4&45.1&31.21& {\color{green}57.3}& 30.3& 34.6& 31.1& 31.0&39.4&40.1&31.3\\
        Fashion~(R-20)&{\color{green}70.5}& 21.1& 22.2& 27.3& 27.8&47.2&30.3&26.2& {\color{green}62.0} & 21.4 & 22.8 & 26.7 & 27.6&42.7&28.5&26.2\\
        SVHN~(L-5)&{\color{green}51.4}& 29.1& 27.5& 30.9& 31.0&41.3&38.9&28.9& {\color{green}51.1} & 29.1 & 28.3 & 30.9 & 31.0&35.4&33.1&28.9\\
        SVHN~(V-16)&{\color{green}54.3} &18.6  &15.5  &17.0  &23.5 &34.7&43.2&15.9& {\color{green}43.3}&18.2&15.6 &16.6 & 23.1&28.9&40.1&15.9\\
        CIFAR-10~(V-16)&{\color{green}63.4} & 27.0 & 17.7 & 30.6 & 42.8&51.9&50.5&30.8&{\color{green}60.7} & 29.8 & 18.3 &30.7 &42.3 &48.5&46.1&30.8 \\
        CIFAR-10~(R-20)&{\color{green}56.7} & 25.6 & 21.5 & 28.8 & 27.2&40.6&42.7&28.9& {\color{green}54.1}& 26.4& 20.7& 28.5& 26.4&35.4&36.9&28.9\\
        \bottomrule
        \end{tabular}

    \caption{Fault Detection Rate of \xxx and baselines. 
    We use green color to highlight the maximum fault detection rate.}
    \label{tab:num_faults}
    \vspace{-0.7cm}
\end{table*}

\subsection{How effective and efficient is \xxx?}\label{sec:eval:selection}
\subsubsection{Fault Detection}
Similar to traditional software testing~\cite{Gligori2015PracticalRT,Zhang2018HybridRT,Legunsen2016AnES}, test case selection tries to find valuable test cases from a large pool of candidate unlabeled test set, which can reduce the cost of manual labeling time once the labeling resource is limited.
For a given selection method, a selected test set that can trigger more faults means it could reveal more defects in the software.
We take \textbf{F}ault \textbf{D}etection \textbf{R}ate~(FDR) as our evaluation metric to measure the effectiveness of \xxx's test case selection method. Specifically, the FDR is defined as follows:
$$FDR(X) = \frac{|X_{wrong}|}{|X|}$$
where $|X|$ denotes the size of the selected test cases, and $|X_{wrong}|$ is the number of test cases misclassified by DNN.

We compared the FDR of \xxx with baselines as shown in \cref{tab:num_faults}, where we compare the FDR of \xxx and our baselines at different test case selection rates~(5\%, 10\%, 15\%, and 20\%).
First, we found that neuron coverage-guided test case selection methods~(i.e., NAC, KMNC, NPC) have lower performance in fault detecting. Sometimes their FDRs are very similar to RS's, which indicates that neuron coverage is not a proper metric to guided test case selection, and this is consistent with previous research~\cite{li2019structural,ismeanful, deepgini,gao2022adaptive}. 

Then we also compare \xxx with DeepGini, the SOTA prioritization technique, which prioritizes test cases based on its Gini value. We can observe although DeepGini has a higher FDR compared with neuron coverage-guided selection, it is still lower than \xxx~(e.g., when selecting 20\% of the test samples, \xxx obtains 20.95\% more FDR than DeepGini in the most advantageous combination Fashion \& ResNet20, even the lowest improvement is an FDR gain of 6.15\% in MNIST \& LeNet1). 
The key reason is that DeepGini prioritizes test cases based on the model output with a high probability of detecting DNN incorrect behavior based on model output, which causes DeepGini only detect test cases where the model has low prediction confidence~(i.e., large Gini values) at the final layer. Test cases where the final output of the model has a high prediction confidence by activating the model's internal sensitive neurons~(e.g., backdoor attack test cases~\cite{Liu2018TrojaningAO,Wang2019NeuralCI}) will not be detected. While \xxx can detect these test cases by analyzing the behaviors of the test cases in the internal neuron. Finally, we compare \xxx with SA metric~(DSA) and active learning baseline~(i.e., ATS), we can observe that \xxx is also better than these selection strategies. Specifically, test cases selected by \xxx have 81.8\% FDR in MNIST\&LeNet1 combination, while DSA only obtains 22.2\% FDR, and ATS only obtains 58.7\% FDR, which indicates that \xxx can select a more valuable unlabeled dataset from the candidate dataset.

Considering the unlabeled candidate dataset used in~\cref{tab:num_faults} was generated by a mixture of seven data augmentation methods, \xxx maybe has a preference for a particular data augmentation method, which will be a threat to validity. We further studied the FDR of our selection strategy under a fixed augmentation unlabeled dataset. We retain a moderate extent of augmentation to stay fully recognizable to the human eye. As listed in \cref{tab:perAug}, \xxx consistently yields a higher FDR compared to random selection~(i.e., +38.5\% improvement when using the data augmentation method of adjusting brightness and +42.7\% using random scaling with Top 5\% test cases selected), indicating that there is no clear preference for a particular augmentation method, thus further ensuring sample selection diversity. From another perspective, since all candidate samples in this experiment are generated using the same augmentation method, it shows that our metric can correctly reflect the intrinsic information of the model and the samples, independent of external specific methods.
We also notice that in blur configuration, the unlabeled dataset will have high FDR in RS~(e.g., 66.3\% FDR when selecting 20\% test cases from the unlabeled dataset) because the CIFAR10 is a low-resolution dataset, making it very destructive to the image contents.

\begin{table}[]
    \centering
    \setlength{\tabcolsep}{2.5pt}
    \begin{tabular}{c|c c c c|c}
    \toprule
         Augmentations& 5\% & 10\% & 15\% & 20\% & Random  \\
    \midrule
         Shift\ (0.15,0.15)& 52.0\%	&47.8\%	&44.9\%	&41.6\% &14.8\%\\
         Rotation\ $(q=10)$& 44.8\%	&40.1\%	&39.1\%	&37.2\% &29.4\%\\
         Scale\ $(r=0.8)$  & 64.4\%	&57.8\%	&54.1\%	&50.2\% &21.7\%\\
         Shear\ $(s=20)$   & 51.2\%	&49.8\%	&46.2\%	&43.4\% &35.4\%\\
         Contrast\ $(\alpha=1.5)$  &48.4\%	&41.2\%	&38.5\%	&35.2\% & 17.1\%\\
         Brightness\ $(\beta=1.5)$ &65.2\%	&56.1\%	&51.7\%	&48.6\% & 26.7\%\\
         Blur\ $(ks=2)$ &81.4\%    &80.6\%	&78.9\% &77.0\% &66.3\%\\
    \bottomrule
    \end{tabular}
    \caption{Fault detection rate~(FDR) with fixed data augmentation method and corresponding parameters. Experiments are conducted with ResNet20 trained on CIFAR-10.}
    \label{tab:perAug}
    \vspace{-0.8cm}
\end{table}

\subsubsection{Fault Detection Diversity}

Insights from traditional software testing found error-inducing inputs are very dense, inspiring DNN testing that detecting a greater variety of errors may also be as important as detecting more errors. We leverage the concept of fault type introduced in~\cite{gao2022adaptive}, which is defined as:
$$
Fault\_Type(x) = (Label(x)^{*} \rightarrow Label(x))
$$
where $Label(x)^{*}$ denotes the ground-truth label, and $Label(x)$ denotes the DNN prediction. For a typical classification dataset with 10 different categories, the number of possible fault types is $10\times9=90$. As the candidate test cases to be selected may not introduce all types of errors, we use the \textit{Fault Type Coverage Rate~(FTCR)}, the proportion of error types introduced by our selected test cases among the error types introduced by all samples, to quantify the ability to select diverse faults.

\begin{figure*}[htbp]
	\centering
	\begin{subfigure}{0.24\linewidth}
		\centering
		\includegraphics[width=1\linewidth]{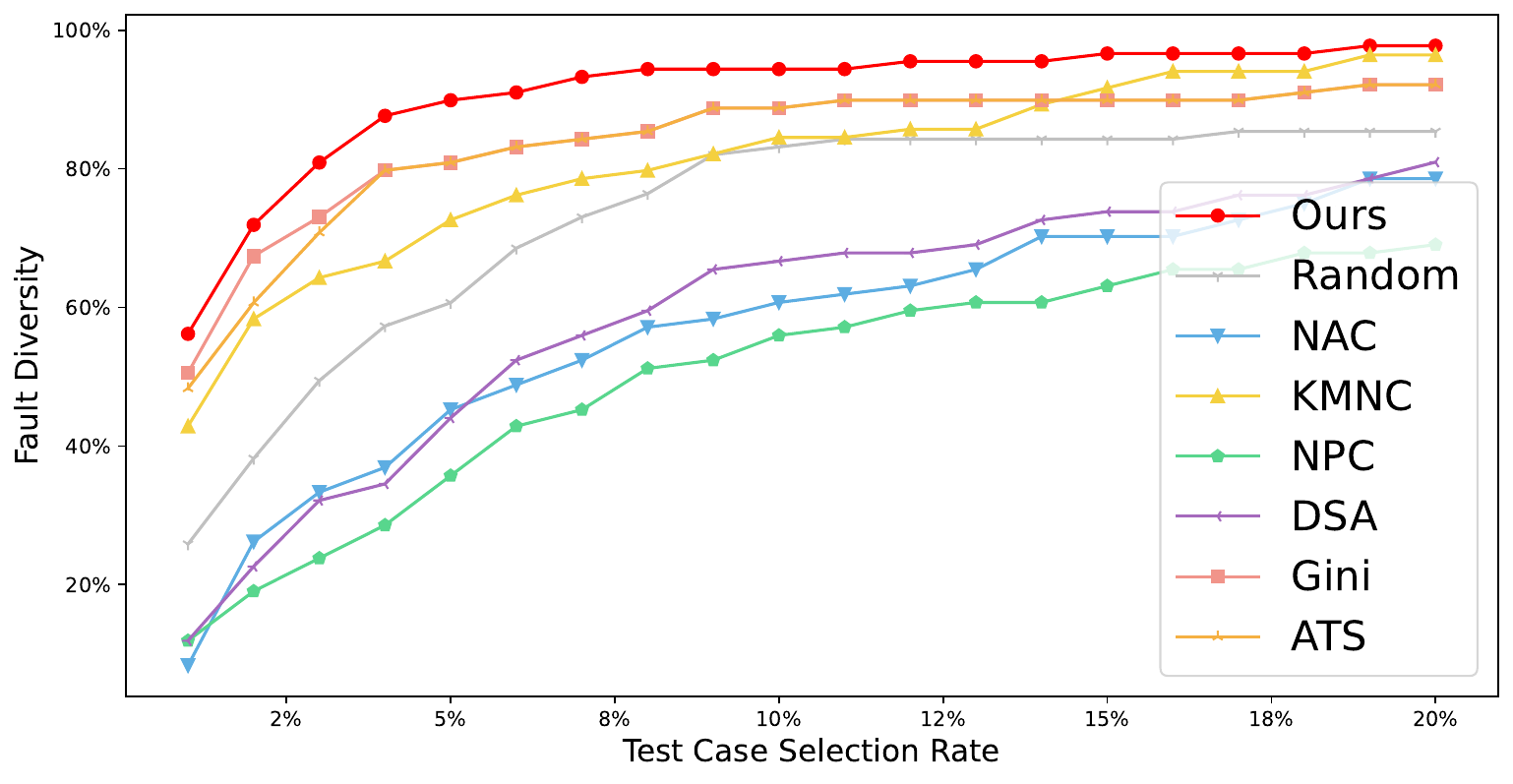}
		\caption{MNIST \& LeNet1}
		\label{chutian3}
	\end{subfigure}
	\centering
	\begin{subfigure}{0.24\linewidth}
		\centering
		\includegraphics[width=1\linewidth]{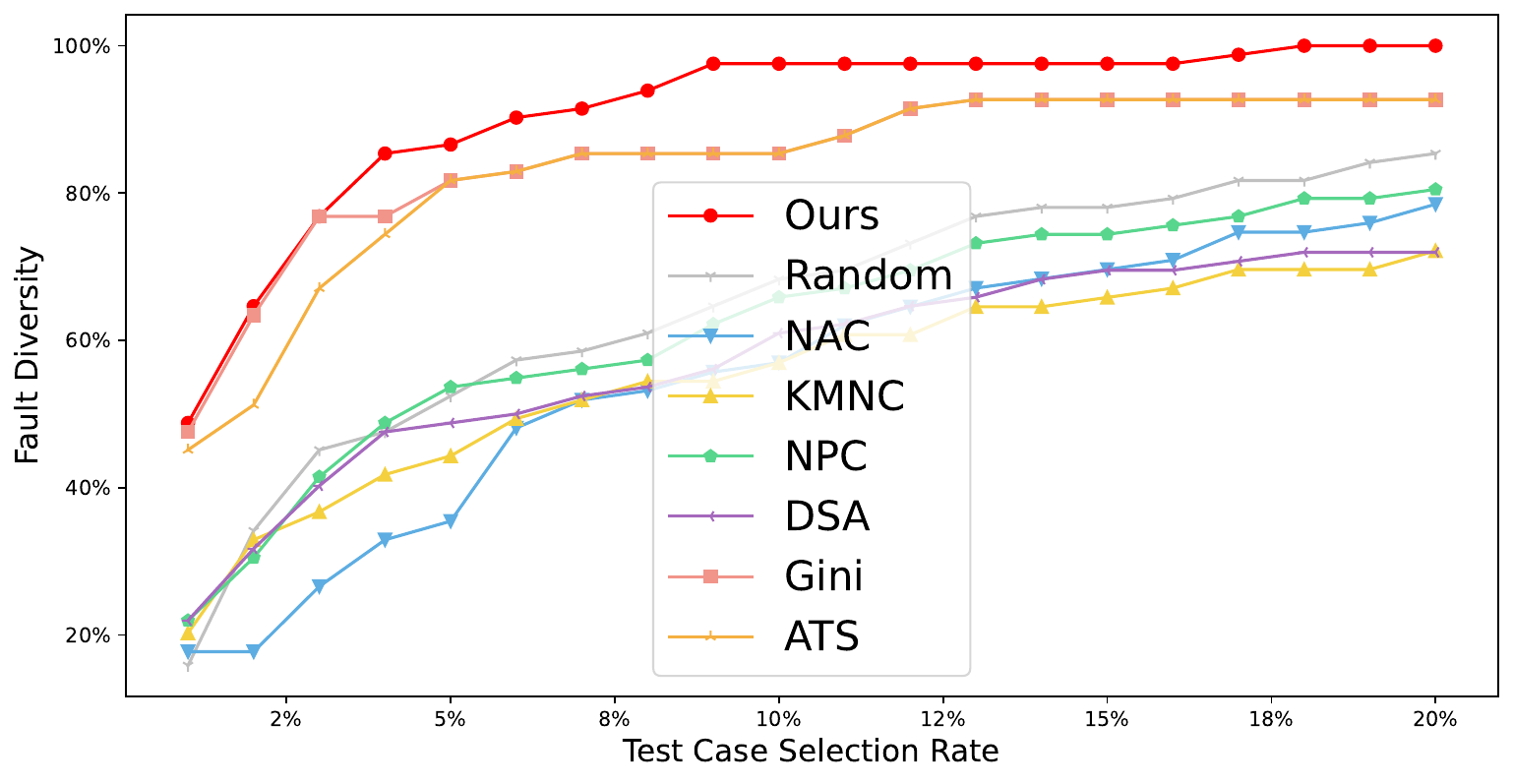}
		\caption{MNIST \& LeNet5}
		\label{chutian3}
	\end{subfigure}
	\centering
	\begin{subfigure}{0.24\linewidth}
		\centering
		\includegraphics[width=1\linewidth]{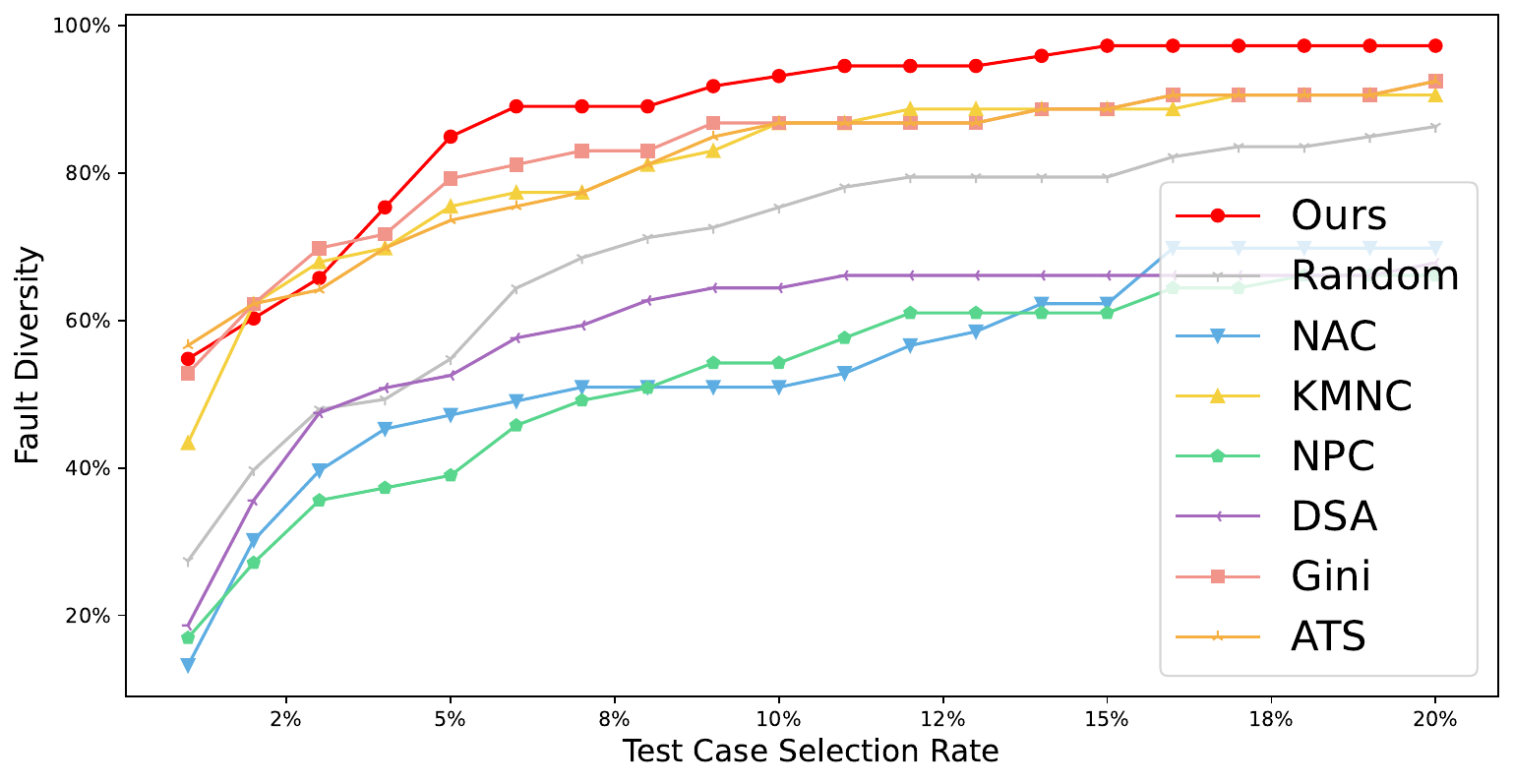}
		\caption{Fashion \& LeNet1}
		\label{chutian3}
	\end{subfigure}
	\begin{subfigure}{0.24\linewidth}
		\centering
		\includegraphics[width=1\linewidth]{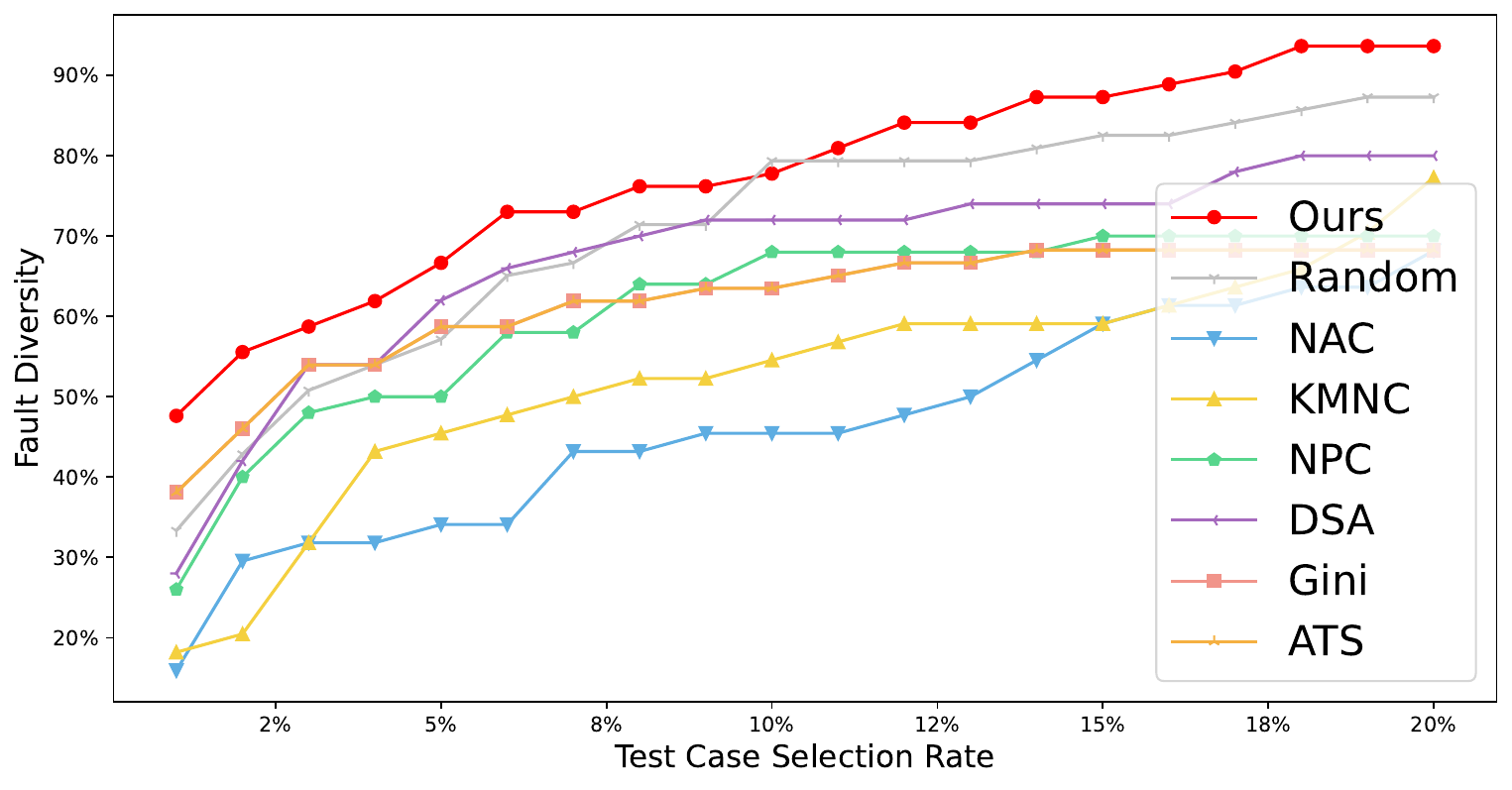}
		\caption{Fashion \& ResNet20}
		\label{chutian3}
	\end{subfigure}
		\begin{subfigure}{0.24\linewidth}
		\centering
		\includegraphics[width=1\linewidth]{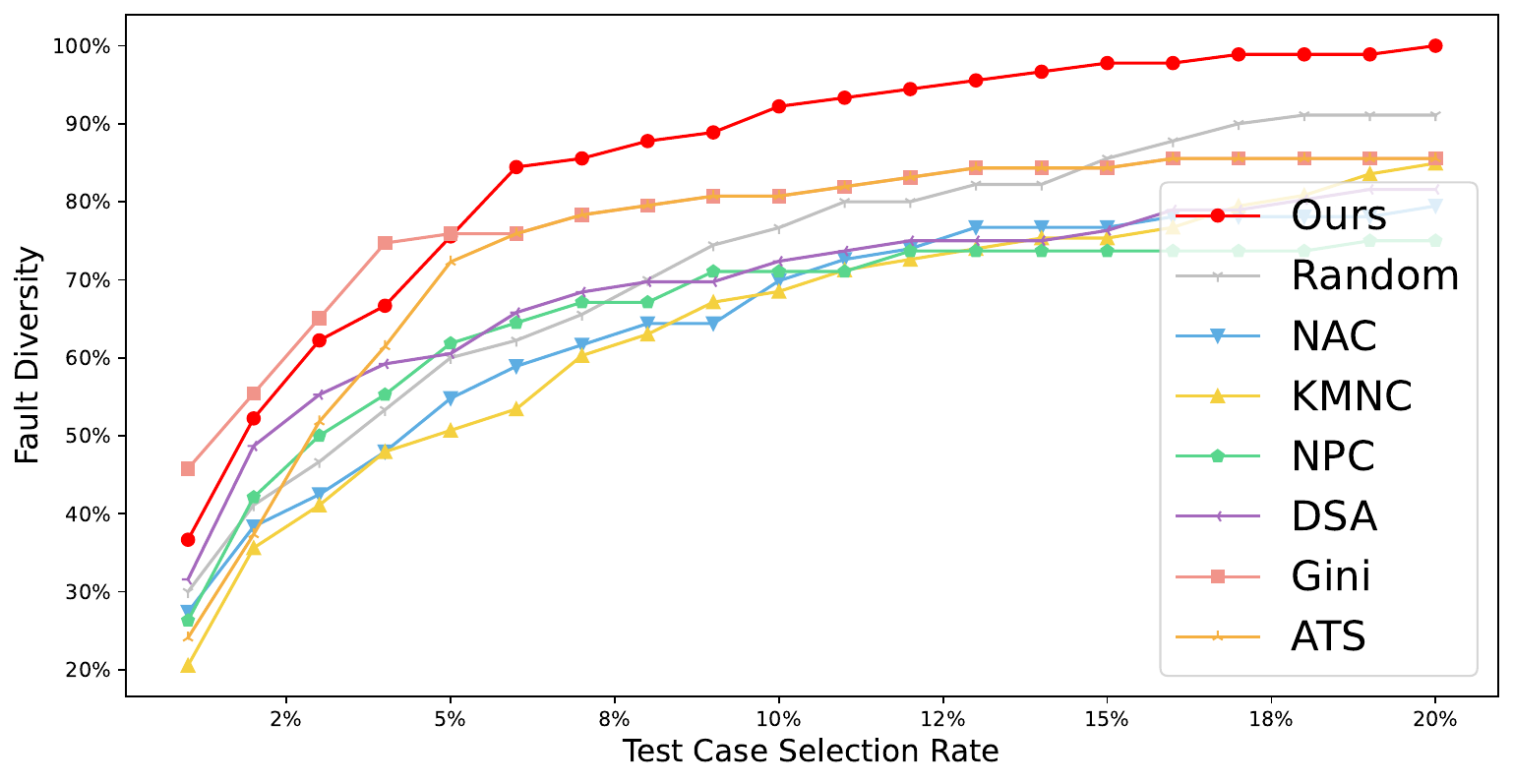}
		\caption{CIFAR10 \& ResNet20}
		\label{chutian3}
	\end{subfigure}
	\centering
	\begin{subfigure}{0.24\linewidth}
		\centering
		\includegraphics[width=1\linewidth]{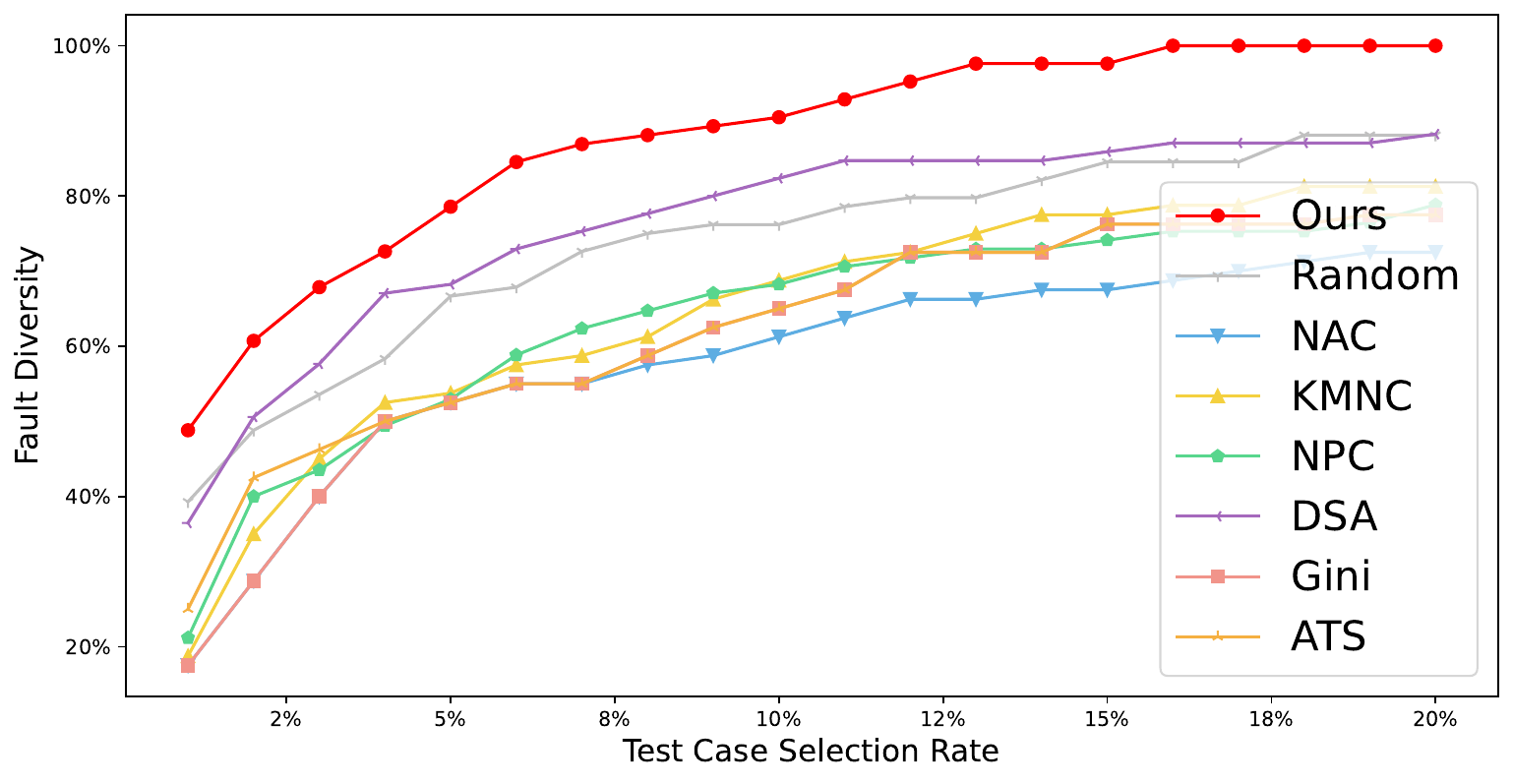}
		\caption{CIFAR10 \& VGG16}
		\label{chutian3}
	\end{subfigure}
	\centering
	\begin{subfigure}{0.24\linewidth}
		\centering
		\includegraphics[width=1\linewidth]{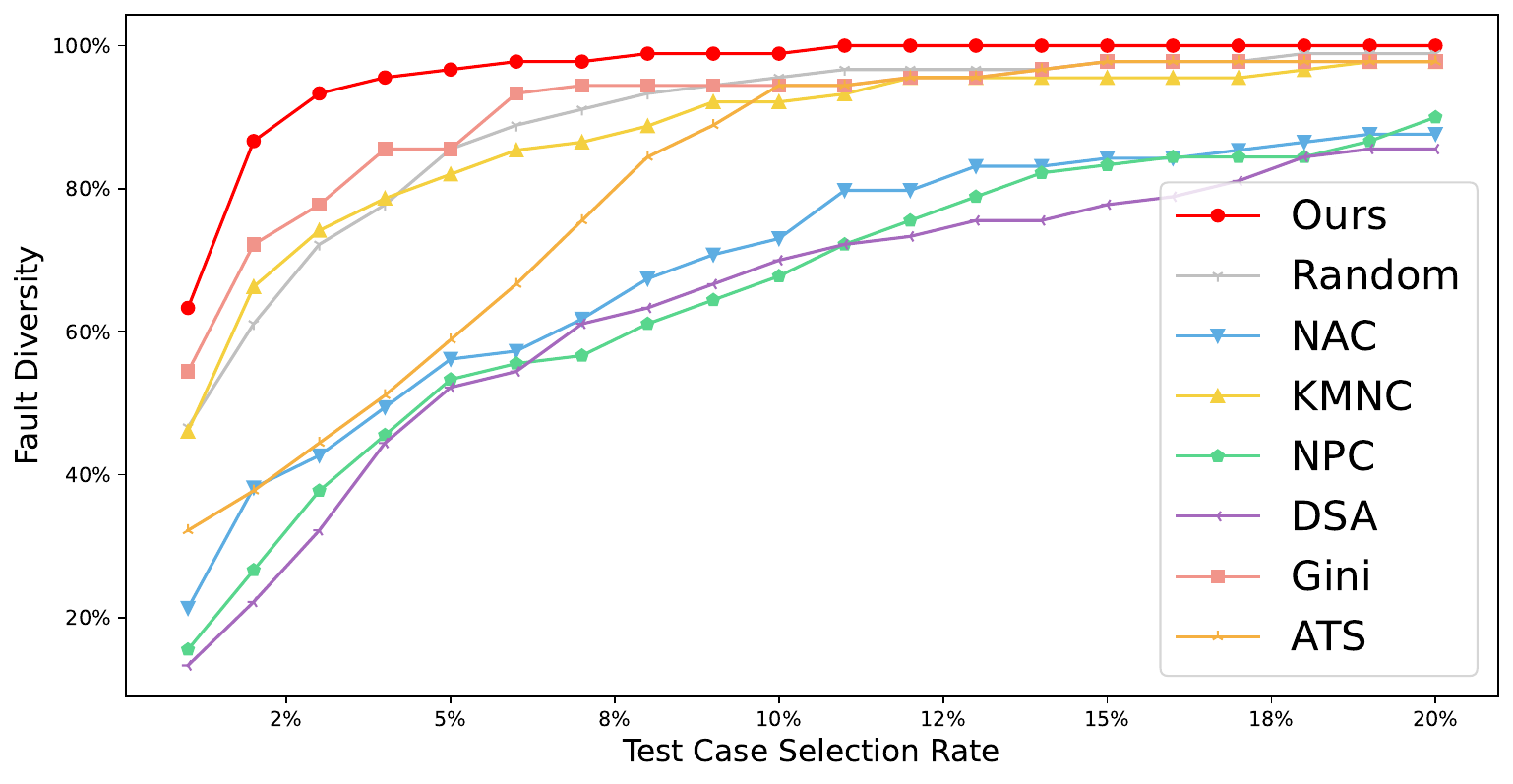}
		\caption{SVHN \& LeNet5}
		\label{chutian3}
	\end{subfigure}
	\begin{subfigure}{0.24\linewidth}
		\centering
		\includegraphics[width=1\linewidth]{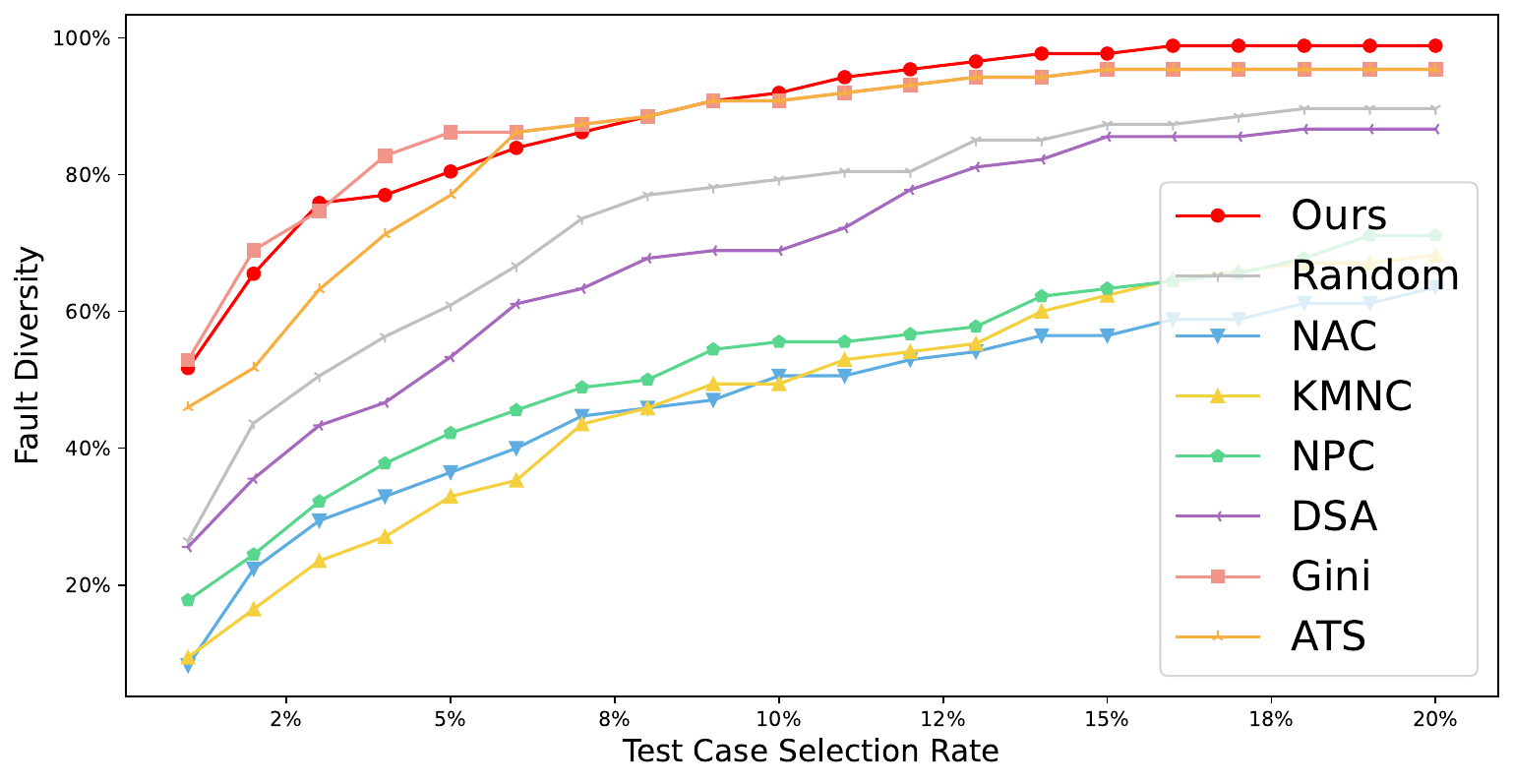}
		\caption{SVHN \& VGG16}
		\label{chutian3}
	\end{subfigure}
	\caption{The cumulative sum of the fault types coverage rate of our and baseline selection methods.}
	\label{fig:diverse}
\end{figure*}

Plots of FTCR with the percentage of selected cases increasing from 1\% to 20\% are shown in~\cref{fig:diverse}. \xxx achieved better fault diversity compared to random selection and baseline methods under all dataset\&model combinations.
Specifically, in selecting more than about 13\% of the candidate samples, the curve of \xxx is always above those of other methods. More importantly, \xxx is the \textbf{only} one of all the methods listed that consistently performs better than random, in terms of fault diversity. Although DeepGini performed comparably to \xxx in MNIST and SVHN experiments, it performed even worse than random selection in Fashion\&ResNet20 and CIFAR10\&VGG16 experiments.
We also calculated the area under the curve to show more accurately the ability of different methods to find diverse errors, as listed in \cref{tab:auc}. It can be seen that existing neuron-coverage metrics are generally weaker than even random selection in terms of the diversity of the corner cases found, while \xxx consistently and significantly outperforms random selection.



\begin{table}[t]
    \setlength{\tabcolsep}{2pt}
    \centering
    \begin{tabular}{l|c c ccc c c c}
        \toprule
        Dataset&\multicolumn{2}{c}{MNIST}& \multicolumn{2}{c}{Fashion}&\multicolumn{2}{c}{CIFAR10}&\multicolumn{2}{c}{SVHN} \\
                Model&L-1&L-5&L-1&R-20&V-16&R-20&L-5&V-16\\ 
         \midrule
        NAC &57.36&55.49&54.12&46.71&58.81&65.53&69.75&47.15\\
        KMNC&81.48&55.86&81.57&52.63&65.39&64.71&88.35&47.98\\
        NPC&50.68&61.29&52.78&63.15&62.07&65.07&65.61&49.84\\
        DSA&59.77&56.46&60.72&70.45&77.03&68.27&63.83&67.02\\
        Gini&85.00&85.33&82.96&62.28&62.23&78.53&91.54&88.47\\
        ATS&84.48&83.98&81.57&62.28&63.48&75.42&80.96&85.69\\
        RS  &74.01&65.37&70.04&71.59&74.15&72.66&90.02&74.62\\
        SCC&{\color{green}88.99}&{\color{green}91.55}&{\color{green}85.11}&{\color{green}72.06}&
        {\color{green}77.26}&{\color{green}86.11}&{\color{green}98.16}&{\color{green}89.08}\\
        \bottomrule
    \end{tabular}
    \caption{When selecting 20\% test cases, the ratio of area under the curve of fault type coverage rate plots. L means LeNet, R means ResNet, and V means VGG.}
    \label{tab:auc}
    \vspace{-0.7cm}
\end{table}

\subsubsection{Optimization Effectiveness}
The goal of DNN testing is to re-train the DNN with selected valuable samples, microscopically fixing the error-induced neurons and macroscopically improving the model's generalization ability. To evaluate the potential superiority of \xxx in optimizing the model, we collect the selected samples with four different ratios~(5\%, 10\%, 15\%, 20\%) and add them to the original training set for further model tuning. The selection strategy employed is consistent with that mentioned in \cref{tab:num_faults}. For each dataset \& model combination, we use the same training hyperparameters~(e.g., epoch, optimizer settings) to perform a fair comparison. Specifically, all models are trained for 40 epochs, with the learning rate initialized to 0.001 and stepped down to one-tenth of the original in the 20th and 30th epochs. We use SGD as the optimizer with a Nesterov momentum of 0.99. Detailed accuracy improvement results are listed in \cref{tab:repairing}.

We can observe that \xxx achieves more significant model improvement than other baseline methods, echoing our discussion for FDR and fault diversity, i.e., choosing more and more diverse test samples that the model would misclassify also helps to optimize the model further and improve its generalization ability. In most cases, the additional performance improvement over the baseline methods is more than 1\%, and in the experiment of Fashion \& LeNet1, the performance improvement is almost twice that of the baseline method. 

\begin{table*}[]
    \centering
    \setlength{\tabcolsep}{2pt}
    \begin{tabular}{l|c|ccccccc|c|ccccccc}
        \toprule
        \multirow{2}{*}{Dataset(DNN)}&\multicolumn{8}{c|}{Select 5\% Test Cases}&\multicolumn{8}{c}{Select 10\% Test Cases}\\
        &\xxx&NAC&KMNC&NPC&DSA&Gini&ATS&RS&\xxx&NAC&KMNC&NPC&DSA&Gini&ATS&RS\\
        \midrule
        MNIST~(L-1)&{\color{green}9.01}&7.07 &7.31 &7.10 &7.10& 7.54&7.41&7.09 &{\color{green}9.04}&7.14 &7.42 &7.14 &7.15&7.91& 7.82&7.14\\
        MNIST~(L-5)&7.44&7.04 &7.24 &7.03 &7.04& {\color{green}7.57}&7.21&7.04 &7.46&7.06 &7.31 &7.08 &7.07& {\color{green}8.21} &7.24&7.06\\
        Fashion~(L-1)&{\color{green}9.43}&4.95 &5.01 &4.98 &4.97&5.31& 5.73& 4.97&{\color{green}9.50}&5.04 &5.11 &5.05 &5.03&5.77& 6.23&5.04\\
        Fashion~(R-20)&{\color{green}6.79}&6.08 &6.10 &6.09 &6.10&6.13&6.15 &6.08 &{\color{green}6.82}&6.14 &6.23 &6.12 &6.16&6.37& 6.42&6.14\\
        SVHN~(L-5)&{\color{green}5.77}&3.86 &3.94 &3.85 &3.86&4.21&4.23 &3.85 &{\color{green}5.85}&4.12 &4.22 &4.13 &4.14&5.21&5.17 &4.13\\
        SVHN~(V-16)&{\color{green}3.35}&2.11 &2.41 &2.12 &2.12&2.57&2.81 &2.11 &{\color{green}3.56}&2.40 &2.51 &2.38 &2.39&2.92&3.03 &2.38\\
        CIFAR-10~(V-16)&{\color{green}2.86}&1.51 &1.61 &1.50 &1.51&2.12&2.05 &1.50 &{\color{green}3.33}&1.82 &1.91 &1.81 &1.82&2.73&2.33 &1.81\\
        CIFAR-10~(R-20)&{\color{green}4.62}&2.35 &2.47 &2.35 &2.34&2.79&2.81 &2.35 &{\color{green}4.73}&2.44 &2.54 &2.43 &2.41&3.02&3.15 &2.43\\
        \midrule
        \multirow{2}{*}{Dataset(DNN)}&\multicolumn{8}{c|}{Select 15\% Test Cases}&\multicolumn{8}{c}{Select 20\% Test Cases}\\
        &\xxx&NAC&KMNC&NPC&DSA&Gini&ATS&RS&\xxx&NAC&KMNC&NPC&DSA&Gini&ATS&RS\\
        \midrule
        MNIST~(L-1)&{\color{green}9.04}&7.20 &7.49 &7.21 &7.20 &8.03&7.93&7.20 &{\color{green}9.07}&7.26 &7.53 &7.26 &7.27&8.07&8.01 &7.26\\
        MNIST~(L-5)&7.46&7.07 &7.37 &7.09 &7.08&{\color{green}8.43} &7.25&7.08 &7.47&7.09 &7.41 &7.10 &7.08&{\color{green}8.62} &7.25&7.08\\
        Fashion~(L-1)&{\color{green}9.53}&5.10 &5.21 &5.09 &5.08&6.01 &6.31&5.09 &{\color{green}9.63}&5.12 &5.24 &5.12 &5.13&6.15&6.47 &5.12\\
        Fashion~(R-20)&{\color{green}6.86}&6.26 &6.31 &6.27 &6.27 &6.52&6.60&6.26 &{\color{green}6.88}&6.31 &6.37 &6.32 &6.31&6.63&6.63 &6.32\\
        SVHN~(L-5)&{\color{green}6.10}&4.38 &4.45 &4.36 &4.37 &5.77&5.48& 4.37 &{\color{green}6.33}&4.61 &4.72 &4.61 &4.59&5.92&5.69 &4.61\\
        SVHN~(V-16)&{\color{green}3.67}&2.57 &2.65 &2.58 &2.57&3.16 &3.17&2.57 &{\color{green}3.72}&2.80 &2.92 &2.82 &2.79&3.32 &3.40&2.81\\
        CIFAR-10~(V-16)&{\color{green}3.84}&2.02 &2.22 &2.04 &2.03&3.21&2.51 &2.04 &{\color{green}4.33} &2.49 &2.57 &2.49 &2.49&3.37 &2.77&2.49\\
        CIFAR-10~(R-20)&{\color{green}4.82}&2.55 &2.71 &2.55 &2.56&3.47 &3.41&2.56 &{\color{green}4.92}&2.71 &2.77 &2.72 &2.72&3.91 &3.76&2.71\\
        \bottomrule
    \end{tabular}
    \caption{Increase in DNN's accuracy~(\%) after repairing the DNN with test cases selected by DLS testing. 
    The greatest increase in a DNN's accuracy among all DLS testing techniques is colored in green.}
    \vspace{-0.5cm}
    \label{tab:repairing}
\end{table*}

\begin{figure*}
	\centering
	\begin{subfigure}{0.24\linewidth}
		\centering
		\includegraphics[width=1\linewidth]{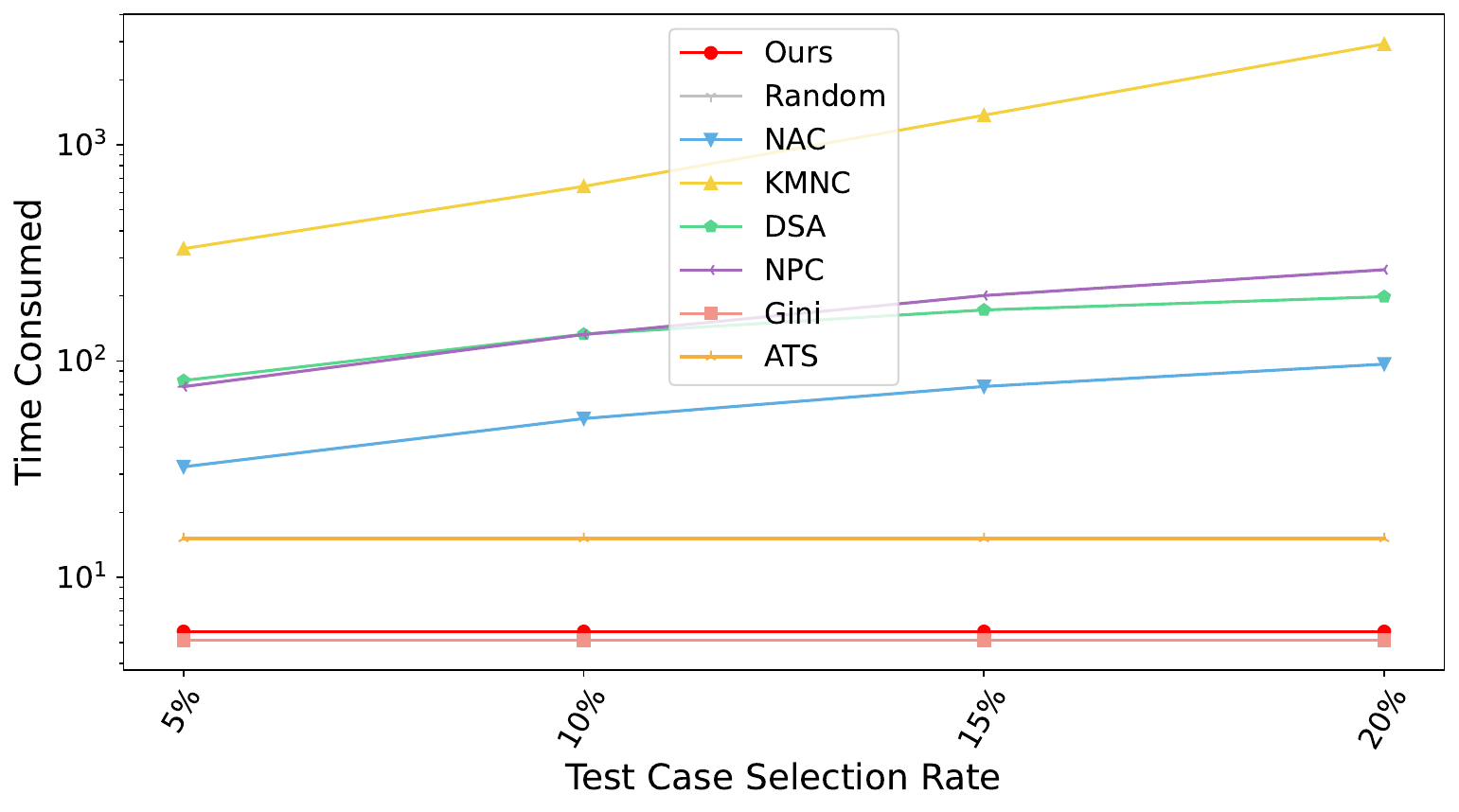}
		\caption{MNIST \& LeNet1}
		\label{chutian3}
	\end{subfigure}
	\centering
	\begin{subfigure}{0.24\linewidth}
		\centering
		\includegraphics[width=1\linewidth]{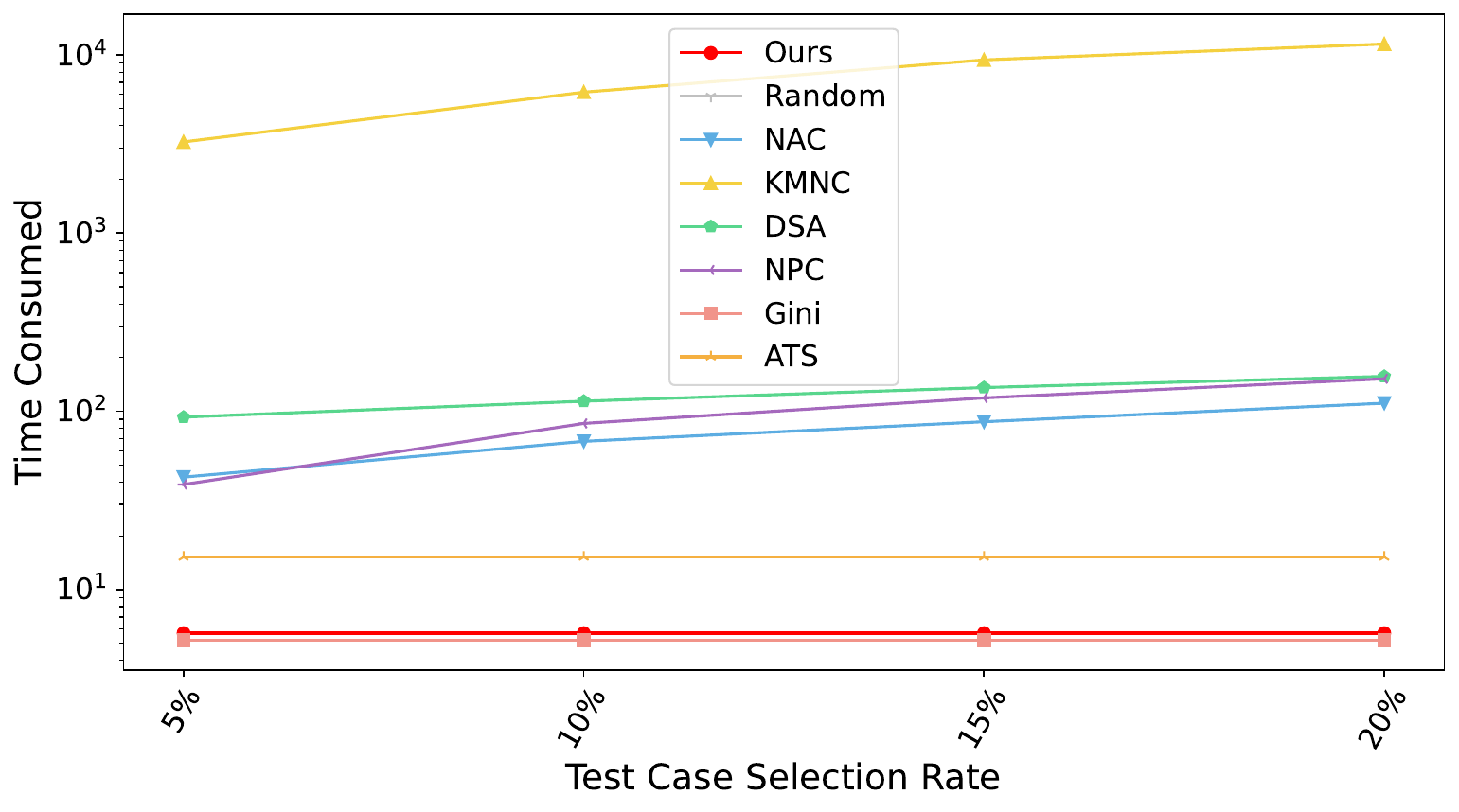}
		\caption{MNIST \& LeNet5}
		\label{chutian3}
	\end{subfigure}
	\centering
	\begin{subfigure}{0.24\linewidth}
		\centering
		\includegraphics[width=1\linewidth]{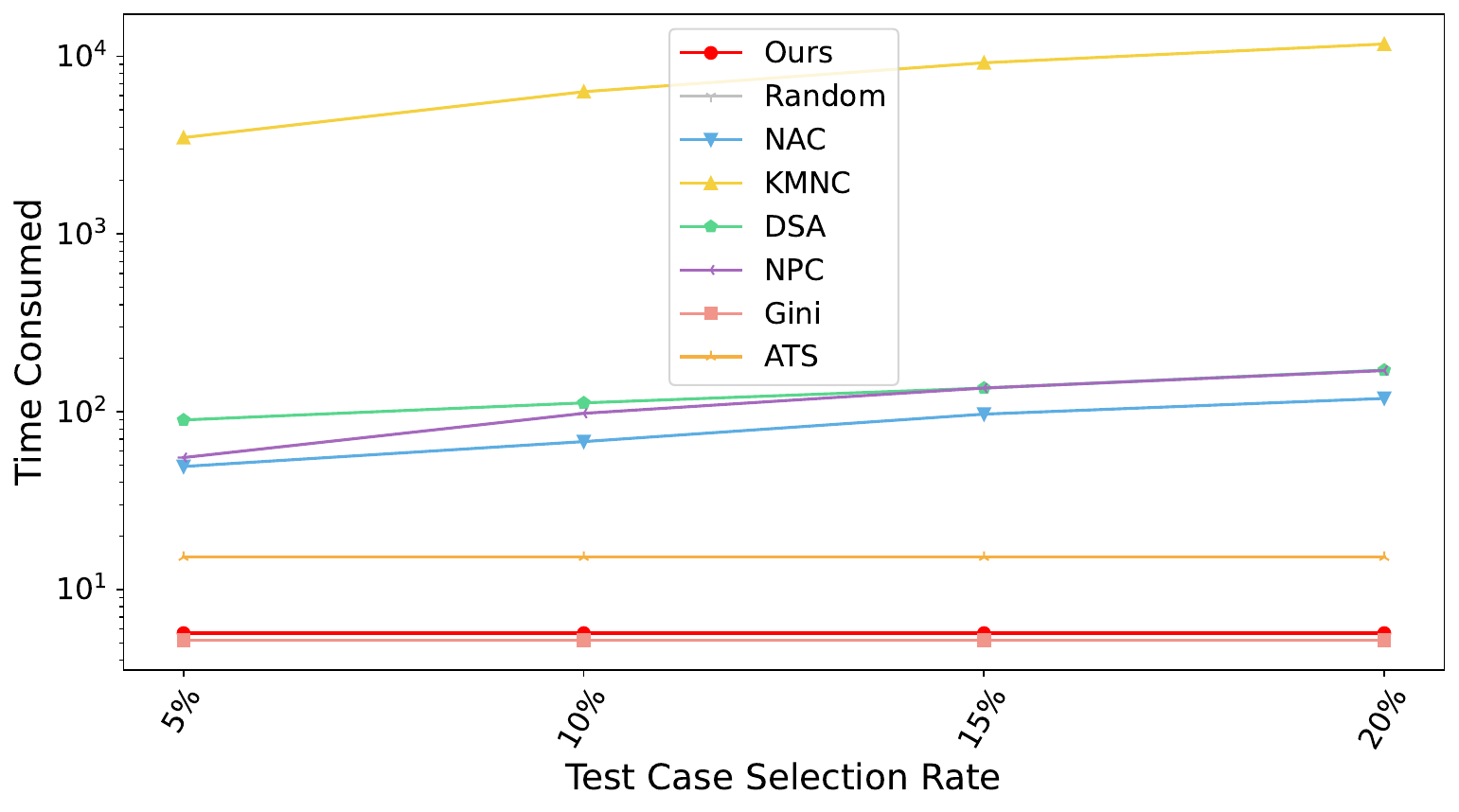}
		\caption{Fashion \& LeNet1}
		\label{chutian3}
	\end{subfigure}
	\begin{subfigure}{0.24\linewidth}
		\centering
		\includegraphics[width=1\linewidth]{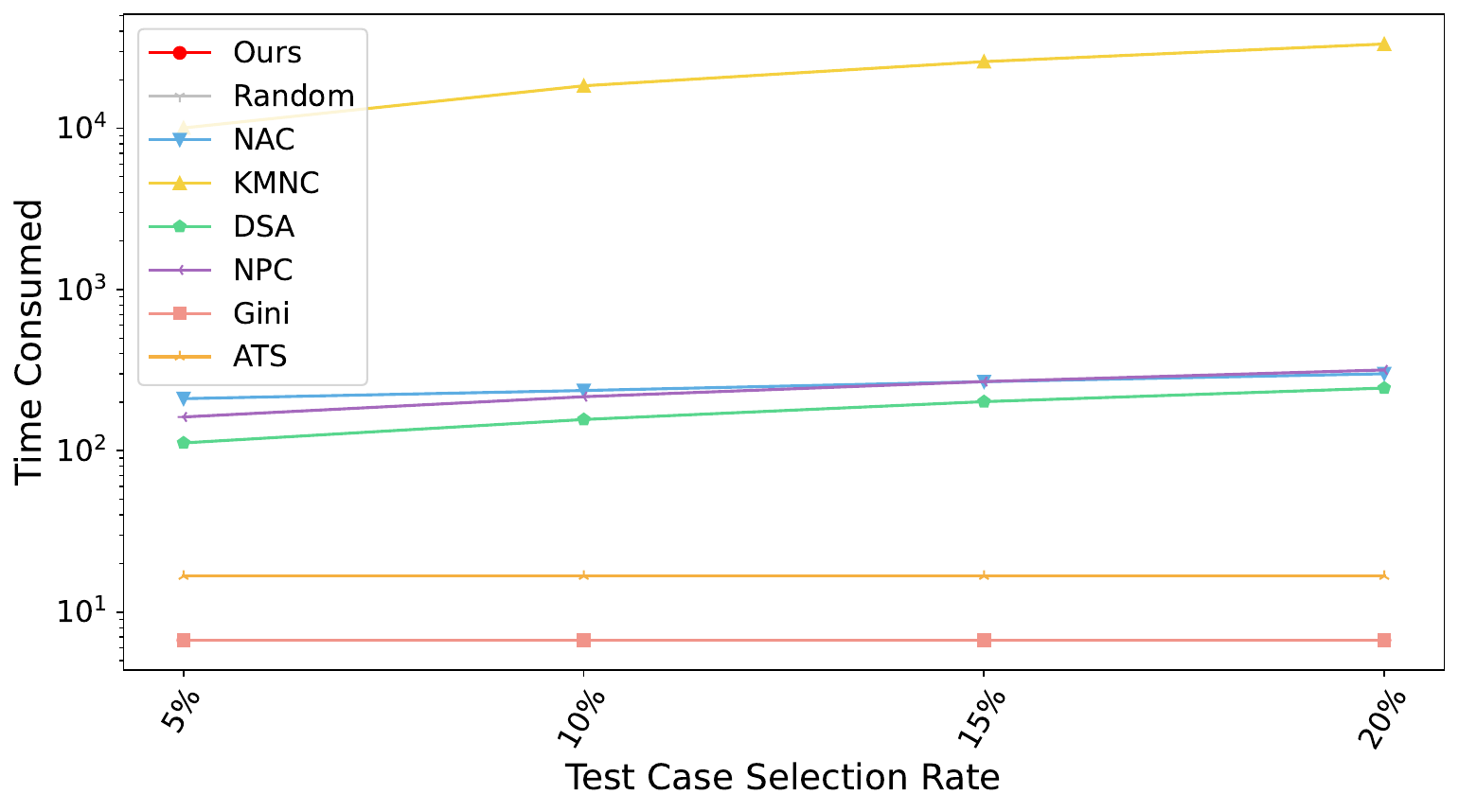}
		\caption{Fashion \& ResNet20}
		\label{chutian3}
	\end{subfigure}
		\begin{subfigure}{0.24\linewidth}
		\centering
		\includegraphics[width=1\linewidth]{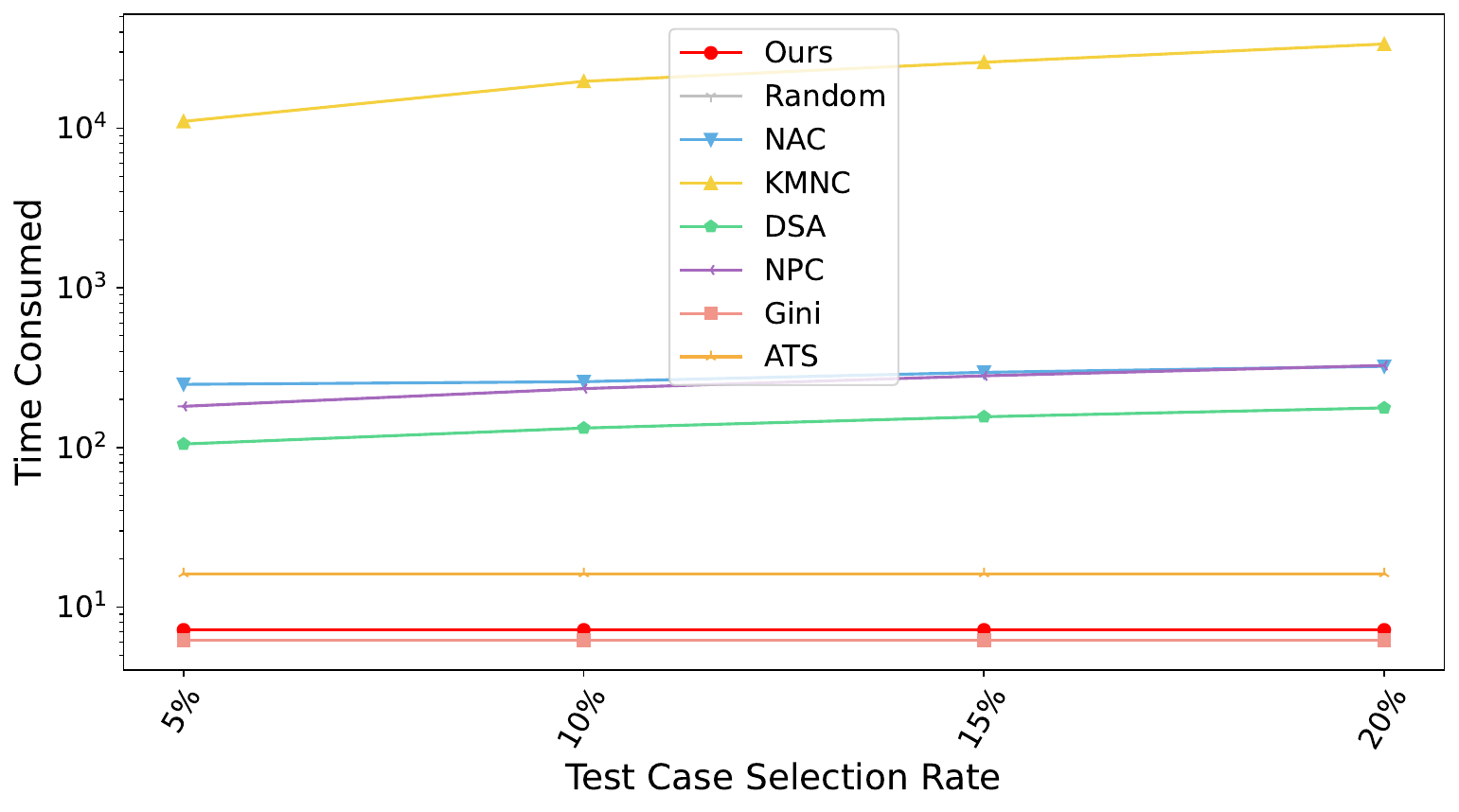}
		\caption{CIFAR10 \& ResNet20}
		\label{chutian3}
	\end{subfigure}
	\centering
	\begin{subfigure}{0.24\linewidth}
		\centering
		\includegraphics[width=1\linewidth]{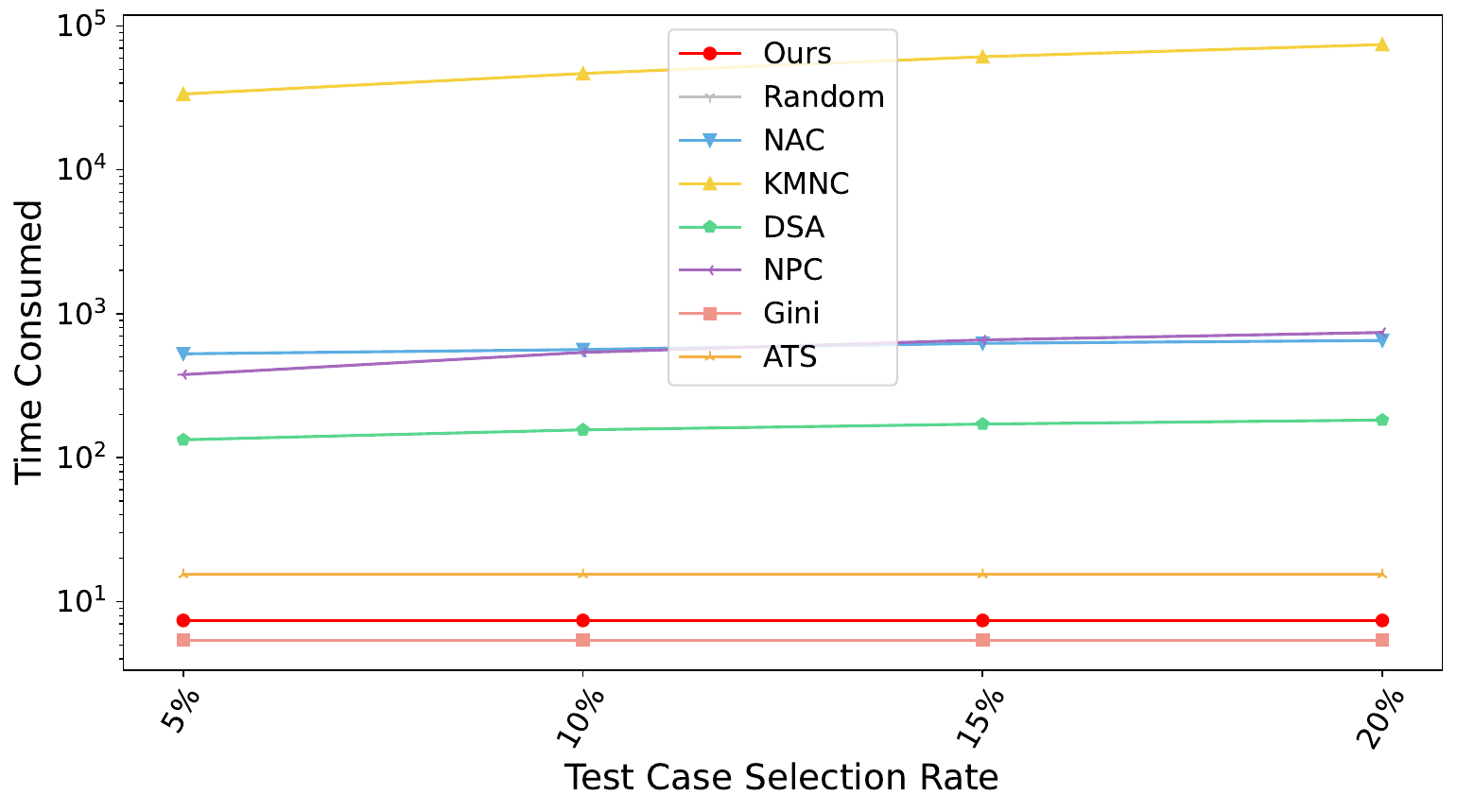}
		\caption{CIFAR10 \& VGG16}
		\label{chutian3}
	\end{subfigure}
	\centering
	\begin{subfigure}{0.24\linewidth}
		\centering
		\includegraphics[width=1\linewidth]{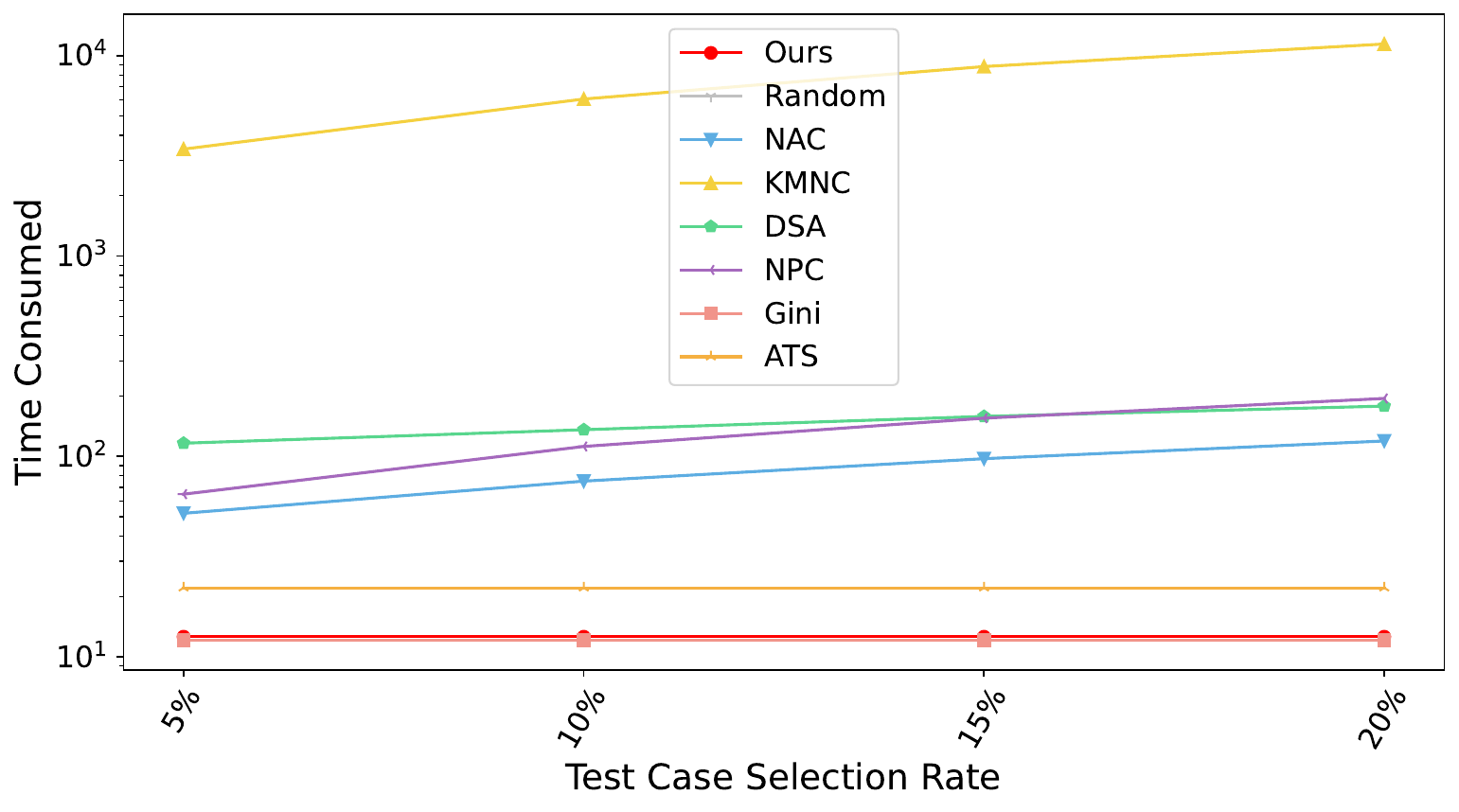}
		\caption{SVHN \& LeNet5}
		\label{chutian3}
	\end{subfigure}
	\begin{subfigure}{0.24\linewidth}
		\centering
		\includegraphics[width=1\linewidth]{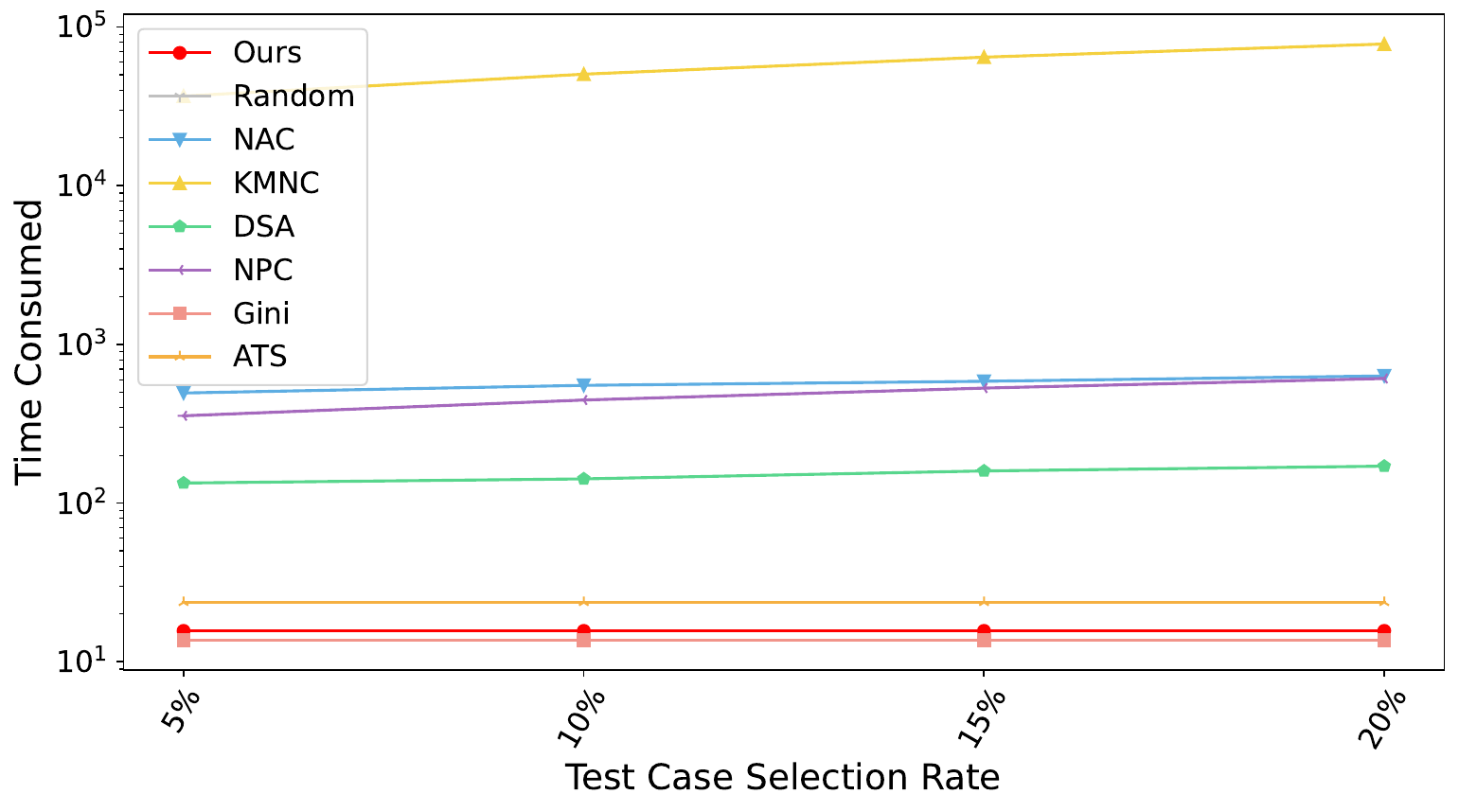}
		\caption{SVHN \& VGG16}
		\label{chutian3}
	\end{subfigure}
	\caption{Overhead on all dataset\&model combinations. The vertical axis is the logarithmic coordinate.}
	\label{fig:time}
\end{figure*}

\subsubsection{Overhead}\label{sec:eval:time}
Test case selection is used to select valuable test cases from a large number of unlabeled candidate datasets, which can reduce the labeling efforts of DNN developers. Therefore, the overhead of the selection method can seriously affect the validity of the method. To evaluate the overhead of \xxx and baselines, we record the time consumed by each selection method when the proportion of selected samples among all candidates is 5\%, 10\%, 15\%, and 20\%, respectively. In this experiment, we take the time consumption for random selection as 0 seconds. Plots are shown in~\cref{fig:time}. 
First, we can observe that neuron coverage guided selection~(i.e., NAC and KMNC) approximately consume  hundreds of seconds on smaller models and close to or over $10^3$ seconds on larger models. The slowest KMNC-based selection generally takes several hours to complete, which is consistent with recent research results~\cite{deepgini,gao2022adaptive}.
In contrast, our approach has a significant order-of-magnitude advantage regarding time overhead, costing only around $10$ seconds with all experimental settings. We can also observe that \xxx is only slower than DeepGini because compared with DeepGini, \xxx will calculate the sensitive neuron's TNSScore, while DeepGini only needs to calculate the final prediction layer's Gini. But we can also notice that calculating TNSScore compared with Gini in the last layer only needs 5s to 10s, compared with the increase of FDR, we believe this overhead is worthwhile. We also notice that ATS will spend more time compared with DeepGini and \xxx, which is because ATS will calculate the extra pattern and fitness score, which cause it slower than \xxx and Gini.

The huge advantage of our algorithm in time is theoretically guaranteed. Typically, similar to DeepGini, we can simply use a quick sort algorithm to sort tests. This algorithm takes $O(nlogn)$ time complexity, where n is the sample size of the candidate test cases. While neuron-coverage metric guided selections are based on a greedy search to select test samples one by one with the time complexity $O(mn^{2})$~\cite{deepgini}, where m is the number of elements (e.g., neurons) to cover and n is the size of unlabeled datasets, the time overhead of \xxx does not grow with the percentage of selected samples as we directly employ quick sorting to sort the sensitivities corresponding to all samples; instead, it is solely correlated with the total number of candidate samples. 

\mybox{Answer to RQ2: \xxx is effective and efficient to select test cases to improve the model performance.
}

\vspace{-0.7cm}

\subsection{How does sensitive neuron size affect \xxx's effectiveness?}\label{sec:eval:size}

The FDR experiment results in \cref{tab:num_faults} are based on Top-10\% most sensitive neurons across all datasets and models, which derives from our empirical studies where we can obtain adequate performance~(FDR) while maintaining sufficient selection efficiency with Top-10\% neurons. However, a fixed percentage of neurons may not generalize well to models, so we also study how the proportion of sensitive neurons tested would affect the estimation of the \textit{value} of a test case and further affect the FDR. Results are listed in \cref{tab:size}. The FDR tends to increase as the percentage of neurons we sample increases, which is consistent with our intuition, as more neurons tend to yield more accurate estimates. However, the testing and repairing overhead will increase once the selected sensitive neuron increases. So we prefer to leave the choice of specific neuron numbers to the DNN developer facing different overhead restrictions given the different accuracy-efficiency trade-off requirements in different scenarios.

\mybox{Answer to RQ3: The ratio of sensitive neuron affect the result of \xxx. Selecting more sensitive neuron can increase the FDR of \xxx.}

\vspace{-0.5cm}

\subsection{How does the selected layer affect \xxx's effectiveness?}\label{sec:eval:layer}
In DNNs, the deeper network layers have a larger perceptual field and contain rich global semantic information, while the shallow layers represent the low-level local information in the data. ~\cref{tab:layer} shows our study on the implications of the choice of model layer on the FDR of \xxx, and we fixed the percentage of sensitive neuron selection at 20\%. Especially for models with more layers~(e.g., ResNet20 and VGG16), the performance gap between shallow and deep layers is substantial~(i.e., 24.6\% FDR difference brought by employing the first and last layer of ResNet20 model trained on CIFAR10). Even for tiny networks like LeNet, leveraging sensitive neurons of the last layer generally yields the highest FDR.

Apart from the stronger representation capability of deeper neurons, which brings a nontrivial contribution to the fault detection performance of our selection method, deeper layers also benefit the efficiency as there are normally fewer neurons contained. The current convolutional network design paradigm entails downsampling the spatial dimension to one-fourth of its original size while doubling the channel dimension, which typically reduces the number of neurons. Taking ResNet20 \& CIFAR10 as an example, the output of the first layer contains 16,384 neurons, while the last contains only 4,096. Locating sensitive neurons from fewer candidates can significantly reduce the time and space consumption of our algorithm. Our layer selection of the last \textbf{encoder}~(illustrated in Fig.\ref{fig:intro}) layer also demonstrates superiority over Gini in most cases as shown in Tab.\ref{tab:num_faults}, who leverage the overall last layer in the DNNs~(i.e., the final fully connected layer in classification models).

\begin{table}[t]
    \centering
    \setlength{\tabcolsep}{2.5pt}
    \begin{tabular}{l|c c c c c}
    \toprule
         \multirow{2}{*}{Dataset~(DNN)}& \multicolumn{5}{c}{Top k\% sensitive neurons}\\
         &1\%&5\%&10\%&20\% &100\%\\
         \hline
         MNIST~(LeNet-1)& 48.9\%	&51.4\%&	53.3\%&	56.7\% & 59.3\%\\
         MNIST~(LeNet-5)& 44.8\%	&50.0\%&	50.9\%&	51.8\% &54.3\%\\
         Fashion~(LeNet-1)& 54.4\%&	57.0\%&	57.4\%&	61.2\% &58.3\%\\
         Fashion~(ResNet-20)& 54.5\%&	60.3\%&	62.0\%&	63.9\% & 66.0\%\\
         SVHN~(LeNet-5)&47.1\%&	50.9\%&	51.2\%&	51.8\% & 52.2\%\\
         SVHN~(VGG-16)&41.4\%&	43.1\%&	43.3\%&	45.7\% & 46.3\%\\
         CIFAR-10~(VGG-16)&55.7\%	&56.6\%&	60.7\%&	60.0\% & 60.9\%\\
         CIFAR-10~(ResNet-20)&52.9\%&	53.9\%&	54.1\%&	54.9\% & 55.1\%\\
         \bottomrule
    \end{tabular}
    \caption{Fault detection rate with the ratio of sensitive neurons~(sampled from the last encoder layer) ranging from 1\% to 100\%. We report the FDR for 20\% of the test cases.}
    \label{tab:size}
\end{table}

\begin{table}[t]
\setlength{\tabcolsep}{2pt}
    \centering
    \begin{tabular}{l|c c c c}
        \toprule
         \multirow{2}{*}{Dataset~(DNN)}& \multicolumn{4}{c}{Sensitive neuron selection layer}\\
         &Layer1&Layer2&Layer-2& Layer-1 \\
         \midrule
         MNIST~(LeNet-1)&56.65&	56.50&	56.70&	60.05\\
         MNIST~(LeNet-5)&51.65&	51.50&	51.80&	53.55\\
         Fashion~(LeNet-1)&57.15	&57.00	&61.20	&60.40\\
         Fashion~(ResNet-20)&44.25	&46.30	&60.65	&63.90\\
         SVHN~(LeNet-5)&47.69	&51.65	&51.79	&56.25\\
         SVHN~(VGG-16)&21.17	&21.65	&42.04	&45.76\\
         CIFAR-10~(VGG-16)&40.45	&40.25	&52.40	&60.00\\
         CIFAR-10~(ResNet-20)&30.25	&33.60	&40.00	&54.85\\
        \bottomrule
    \end{tabular}
    \caption{\xxx's \fdr~(\%) with different model layers where sensitive neurons are sampled from. \textit{Layer-1} denotes the last layer of the encoder part of the network.}
    \label{tab:layer}
\end{table}

\mybox{Answer to RQ4: The selected layer will affect the result of \xxx. Since the last encoder layer is responsible to understand semantically high-level features, we argue it is a suitable choice for \xxx.}

\subsection{Threats to Validity}

\paragraph{\textbf{Test subject selection.}} The selection of evaluation subjects (i.e., datasets and DNN models) could be threat to validity. We try to counter this by using four commonly studied datasets; for DNN models, we use four well-known pre-trained DNN-based models  of different sizes and complexity ranging from 3,350 neurons up to more than 35,749,834 neurons.

\paragraph{\textbf{Data simulation.}} Another threat to validity comes from the augmented test input generation. To generate test cases, we choose seven widely-used benign mutations as our baselines to simulate faults from different sources and granularity. Although these data simulations are very similar to virtual environment noise, it is impossible to guarantee that the distribution of the real unseen input is the same as our simulation. Additional experiments based on real unseen inputs need to be conducted in future work.

\paragraph{\textbf{Parameters settings.}} The parameter settings in baselines could also be a threat to validity. To compare with neuron coverage-guided and prioritization test set selection methods, we reproduced existing selection methods for DNN, which may include parameters. By fine-tuning the parameter settings, the selected data could be different. To alleviate the potential bias, we follow the authors’ suggested settings or employ the default settings of the original papers.
\section{Related Work}
This section discusses the related work in two groups: prioritization techniques and coverage-guided testing.

\subsection{Prioritization Techniques}

Test prioritization tries to find the ideal ordering of test cases so that software testers or developers can obtain maximum benefit within a limited budget. The idea was first mentioned by ~\citet{Wong1995EffectOT}, and then the technique was proposed by ~\citet{Rothermel1996AnalyzingRT,Harrold1999TestingES} in a more general context. We observe that such an idea from the area of software engineering can significantly reduce the effort of labeling for deep learning systems. This is mainly because a deep learning system usually has a large number of unlabeled tests, but developers only have limited time for labeling.

Inspired by prioritization techniques in conventional software testing, some prioritization techniques~\cite{deepgini,CES} are proposed to prioritize unlabeled datasets in deep learning testing, which can reduce the labeling time by only labeling partially valuable datasets ~(i.e., test cases).~\citet{deepgini} propose DeepGini, which prioritize test case based on their Gini value in the final layer. ~\citet{CES} propose CES, which prioritizes the test case based on the test case's Cross Entropy-based Sampling in the final layer's output. However, these prioritization techniques are focused on the model's final layer's output and ignore utilizing internal neuron information, which causes they can not to detect test cases that can induce internal neurons into an error and then cause the model to have incorrect behaviors even though the final layer output has high confidence. In contrast, our method utilizes internal neuron information by proposing the test case's neuron sensitivity score to reveal how a test case affects internal neuron behaviors, which causes our method to become more effective and efficient.


\subsection{Coverage-Guided Testing}
\citet{pei2017deepxplore} proposes the first white-box coverage criteria (i.e., NAC), which calculates the percentage of activated neurons. DeepGauge~\cite{ma2018deepgauge} then extends NAC and proposes a set of more fine-grained coverage criteria by considering the distribution of neuron outputs from the training data.
Inspired by the coverage criteria in traditional software testing, some coverage metrics~\cite{Deepcover,Ma2019DeepCTTC,Ma2018DeepMutationMT} are proposed. DeepCover~\cite{Deepcover} proposes the MC/DC coverage of DNNs based on the dependence between neurons in adjacent layers. To explore how inputs affect neuron internal decision logic flow,~\citet{Xie2022NPCNP} propose NPC to txplore the neuron path coverage.
DeepCT~\cite{Ma2019DeepCTTC} adopts the combinatorial testing idea and proposes a coverage metric that considers the combination of different neurons at each layer. DeepMutation~\cite{Ma2018DeepMutationMT} adopts the mutation testing into DL testing and proposes a set of operators to generate mutants of the DNN. Furthermore, DeepConcolic~\cite{Sun2018ConcolicTF} analyzed the limitation of existing coverage criteria and proposed a more fine-grained coverage metric that considers the relationships between two adjacent layers and the combinations of values of neurons at each layer.
Our study demonstrated that using these coverage criteria, coverage-based test prioritization is not effective and efficient~(\cref{sec:eval:selection}), which is usually time-consuming and highly expensive. Sometimes, its effectiveness is even worse than random prioritization. Instead, our approach uses a simple metric by analyzing internal neuron sensitivity to become more effective and efficient.

\section{Conclusion}
In this paper, we propose \xxx, a neuron sensitivity-guided test case selection method, and we also propose a sensitive neuron identifier to detect sensitive neurons in the model. The experimental results of the paper show that \xxx can efficiently and quickly find more valuable test cases, which can be used to improve the quality of the model. \xxx combines the advantages of existing neuron coverage criteria and prioritization techniques (i.e., allows the selected test cases to find neuron's corner cases), and these test cases also have higher confidence triggering the model into error. These test cases can be used to better improve the quality of the model. \xxx's source code and all evaluation results are available at GitHub Page~\cite{sourcecode}.

\bibliographystyle{IEEEtranN}
\bibliography{IEEEfull}

\end{document}